\documentclass[runningheads]{llncs}
\usepackage[T1]{fontenc}
\usepackage{graphicx}
\usepackage{hyperref}
\usepackage{color}

\usepackage{amsmath,amssymb}
\usepackage{url}

\usepackage{authblk} % TODO

\newcommand{\refequation}[1]{Equation~\eqref{#1}}
\newcommand{\reffigure}[1]{\figurename~\ref{#1}}

\newcommand{\refsubsection}[1]{Subsection~\ref{#1}}

\usepackage[skins,breakable,xparse]{tcolorbox}
\tikzset{
  dirtree/.style={
    grow via three points={one child at (0.8,-0.7) and two children at (0.8,-0.7) and (0.8,-1.45)}, 
    edge from parent path={($(\tikzparentnode\tikzparentanchor)+(.4cm,0cm)$) |- (\tikzchildnode\tikzchildanchor)}, growth parent anchor=west, parent anchor=south west},
}

\usepackage{listingsutf8}

\lstdefinelanguage{ludii}{
  keywords={aPI,card,component,die,piece,domino,path,tile,board,boardless,surakartaBoard,hashiBoard,puzzleBoard,track,container,deck,dice,hand,equipment,item,dominoes,hints,map,regions,all,allDiceEqual,allDiceUsed,allPassed,allType,allClaimed,baseBooleanFunction,booleanConstant,booleanFunction,can,canMove,canType,all,allDifferent,allPuzzleType,forAll,isUnique,is,isPuzzleGraphType,isPuzzleRegionResultType,isPuzzleSimpleType,isCount,isSum,allConnected,isProduct,isShape,sameParity,isSolved,isThreatened,isWithin,isBlocked,isConnected,isCrossing,isLastFrom,isLastTo,isIn,isInvisible,isMasked,isVisible,isAnyDie,isEven,isFlat,isOdd,isPipsMatch,isSidesMatch,isVisited,is,isComponentType,isConnectType,isEdgeType,isGraphType,isIndexPlayerType,isIntegerType,isInType,isLineType,isLoopType,isPathType,isPlayerType,isRegularGraphType,isRelationType,isSimpleType,isSiteType,isStringType,isTargetType,isTreeType,isLine,isLoop,isPath,isEnemy,isFriend,isMover,isNext,isPrev,isTriggered,isRegularGraph,isRelated,isCycle,isFull,isPending,isEmpty,isOccupied,isDecided,isProposed,isTarget,isCaterpillerTree,isSpanningTree,isTree,isTreeCentre,isPlanarGraph,isPlanarGraph2,and,equals,ge,gt,if,le,lt,not,notEqual,or,xor,no,noMoves,noType,was,wasPass,wasType,directions,directionsFunction,if,baseFloatFunction,floatConstant,floatFunction,baseGraphFunction,basis,brick,brickShapeType,diamondOrPrismOnBrick,spiralOnBrick,squareOrRectangleOnBrick,celtic,customOnHex,diamondOnHex,hex,hexagonOnHex,hexShapeType,rectangleOnHex,starOnHex,triangleOnHex,customOnMesh,mesh,morris,quadhex,customOnSquare,diagonalsType,diamondOnSquare,rectangleOnSquare,square,squareShapeType,tiling,tiling31212,tiling333333_33434,tiling33336,customOn33344,tiling33344,tiling33434,customOn3464,diamondOn3464,hexagonOn3464,parallelogramOn3464,rectangleOn3464,starOn3464,tiling3464,tiling3464ShapeType,triangleOn3464,customOn3636,tiling3636,tiling4612,customOn488,squareOrRectangleOn488,tiling488,tilingType,customOnTri,diamondOnTri,hexagonOnTri,rectangleOnTri,starOnTri,tri,triangleOnTri,triShapeType,circle,rectangle,repeat,shape,shapeTypeStar,spiral,wedge,graphFunction,add,clip,complete,dual,hole,intersect,keep,makeFaces,merge,remove,renumber,rotate,scale,shift,skew,splitCrossings,subdivide,trim,union,baseIntArrayFunction,intArrayConstant,intArrayFunction,difference,degrees,groupSizes,rotations,baseIntFunction,ahead,centrePoint,column,coord,cost,handSite,id,layer,mapEntry,phase,row,where,card,cardSimpleType,cardSiteType,cardTrumpSuit,cardRank,cardSuit,cardTrumpRank,cardTrumpValue,groupProduct,tree,between,edge,from,hint,level,site,to,track,var,countPieces,countPips,count,countComponentType,countSimpleType,countSiteType,countSpaceType,countStepsType,countActive,countCells,countColumns,countEdges,countMoves,countMovesThisTurn,countPhases,countPlayers,countRows,countTrials,countTurns,countVertices,countAdjacent,countDiagonal,countNeighbours,countNumber,countOff,countOrthogonal,countSites,countGroups,countLiberties,countSteps,countDegreeProduct,countLeaves,countNonLeaves,countNonLeavesDegree,countTreeDepth,countTreeSize,face,pips,intConstant,intFunction,last,lastFrom,lastTo,lastType,matchScore,abs,add,div,if,max,min,mod,mul,pow,sub,sizeGroup,sizeTerritory,sizeStack,size,sizeGroupType,sizeSiteType,sizeTerritoryType,topLevel,amount,counter,mover,next,previous,score,state,what,who,pathExtent,trackSiteMove,trackSiteEndTrack,trackSite,trackSiteMoveType,trackSiteType,valuePiece,valuePlayer,valuePending,value,valueComponentType,valuePlayerType,valueSimpleType,tree,triangleGroupCount,baseRangeFunction,exact,max,min,range,rangeFunction,baseRegionFunction,forEach,difference,expand,if,intersection,union,regionConstant,regionFunction,sitesAround,sitesCoords,sitesCrossing,sitesCustom,sitesDirection,sitesDistance,sitesAngled,sitesAxial,sitesHorizontal,sitesSlash,sitesSlosh,sitesVertical,sitesGroup,sitesIncident,sitesCell,sitesColumn,sitesEdge,sitesEmpty,sitesPhase,sitesRow,sitesState,sitesLineOfSight,sitesOccupied,sitesEquipmentRegion,sitesHand,sitesInvisible,sitesMasked,sitesStart,sitesTrack,sitesVisible,sitesWinning,sitesSide,sitesBoard,sitesBottom,sitesCentre,sitesConcaveCorners,sitesConvexCorners,sitesCorners,sitesHint,sitesInner,sitesLastFrom,sitesLastTo,sitesLeft,sitesLineOfPlay,sitesMajor,sitesMinor,sitesOuter,sitesPending,sitesPlayable,sitesRight,sitesToClear,sitesTop,sites,sitesAroundType,sitesCrossingType,sitesDirectionType,sitesDistanceType,sitesEdgeType,sitesGroupType,sitesIncidentType,sitesIndexType,sitesLineOfSightType,sitesOccupiedType,sitesPlayerType,sitesSideType,sitesSimpleType,sitesTrackType,sitesWalk,borderSites,game,games,match,subgame,mode,player,players,baseEndRule,byScore,end,endRule,forEach,if,result,automove,meta,metaRule,noRepeat,swap,nextPhase,phase,baseMoves,decisions,move,moveBetType,moveFromToType,moveHopType,moveLeapType,moveMessageType,movePromoteType,moveRemoveType,moveSelectType,moveSetType,moveShootType,moveSimpleType,moveSiteType,moveSlideType,moveStepType,add,apply,bet,claim,effects,fromTo,hop,leap,note,attract,custodial,deal,directionCapture,enclosed,flip,allCombinations,and,append,if,or,push,roll,setDirection,setCounter,setVar,setNextPlayer,setPending,setScore,setValue,set,setDirectionType,setNextPlayerType,setPendingType,setPlayerType,setRegionType,setSiteType,setTrumpType,setValueType,setCount,setState,setInvisible,setMasked,setVisible,setTrumpSuit,sow,addScore,moveAgain,rememberState,shiftPlayers,swapPlayers,swapPieces,swap,swapPlayersType,swapSitesType,surrounded,takeControl,takeDomino,take,takeControlType,takeSimpleType,then,trigger,pass,playCard,promote,propose,remove,avoidStoredState,do,firstMoveOnTrack,maxDistance,max,maxDistanceType,maxMovesType,maxCaptures,maxMoves,priority,satisfy,select,shoot,slide,step,vote,forEachDie,forEachDirection,forEach,forEachDieType,forEachDirectionType,forEachPieceType,forEachPlayerType,forEachSiteType,forEachPiece,forEachPlayer,forEachSite,moves,play,rule,rules,deal,set,placeItem,place,placeRandomType,placeStackType,placeRandom,placeCustomStack,placeMonotonousStack,setAmount,setScore,setTeam,set,setStartGraphType,setStartPlayersType,setStartPlayerType,setStartSitesType,allInvisible,setCost,setCount,setPhase,setSite,split,splitType,start,startRule,basisType,landmarkType,puzzleElementType,regionTypeDynamic,regionTypeStatic,relationType,shapeType,siteType,stepType,tilingBoardlessType,trackType,cardType,dealableType,suitType,dummy,modeType,repetitionType,resultType,roleType,whenType,gameType,noType,absoluteDirection,compassDirection,direction,directionType,directionUniqueName,relativeDirection,rotationalDirection,spatialDirection,dummy,score,card,hint,region,values,edge,face,graph,graphElement,itemScore,measureGraph,perimeter,poly,properties,radial,radials,step,steps,trajectories,vertex,count,pair,between,flips,from,to,what,who,ai,aIItem,features,featureSet,heuristics,centreProximity,cornerProximity,currentMoverHeuristic,heuristicTerm,lineCompletionHeuristic,material,mobilitySimple,ownRegionsCount,playerRegionsProximity,playerSiteMapCount,regionProximity,score,sidesProximity,divNumBoardSites,divNumInitPlacement,heuristicTransformation,logisticFunction,tanh,bestAgent,pair,board,boardBooleanType,boardColourType,boardShapeType,boardStyleThicknessType,boardStyleType,boardCheckered,boardColour,boardShape,boardStyle,boardStyleThickness,graphics,graphicsItem,noAnimation,noBoard,noCurves,noHandScale,noMaskedColour,no,noBooleanType,adversarialPuzzle,stackType,suitRanking,pieceColour,pieceColourFromState,pieceFamilies,pieceBackground,pieceForeground,pieceAddStateToName,pieceExtendName,pieceRename,piece,pieceColourFromStateType,pieceColourType,pieceFamiliesType,pieceGroundType,pieceNameType,pieceReflectType,pieceScaleByType,pieceScaleType,pieceStyleType,pieceReflect,pieceScale,pieceScaleByValue,pieceStyle,playerColour,player,playerColourType,regionColour,region,regionColourType,showCost,showPits,showPlayerHoles,showRegionOwner,showCheck,showPieceState,showPieceValue,showEdges,showScore,show,showBooleanType,showCheckType,showComponentDataType,showComponentType,showEdgeType,showScoreType,showSiteDataType,showSiteType,showSymbolType,showSitesAsHoles,showSitesShape,showSymbol,boardGraphicsType,colour,colourRoutines,userColourType,componentStyleType,containerStyleType,controllerType,edgeInfoGUI,edgeType,lineStyle,metadataFunctions,metadataImageInfo,pieceStackType,valueLocationType,whenScoreType,aliases,author,classification,credit,date,description,origin,publisher,rules,source,version,info,infoItem,metadata,metadataItem},
  basewidth  = {.6em,0.6em},
  keywordstyle=\color{mblue}\bfseries,
  ndkeywords={Off,End,Undefined,DiceUsed,DiceEqual,Passed,Move,Different,Unique,Count,Sum,Solved,Threatened,Within,Connected,Blocked,Crossing,LastFrom,LastTo,Visible,Masked,Invisible,Odd,Even,Visited,SidesMatch,PipsMatch,Flat,AnyDie,In,Line,Loop,Path,Mover,Next,Prev,Friend,Enemy,Triggered,RegularGraph,Related,Cycle,Pending,Full,Empty,Occupied,Proposed,Decided,Target,Tree,SpanningTree,CaterpillerTree,TreeCentre,Moves,Pass,Square,Rectangle,Diamond,Prism,Spiral,Limping,NoShape,Square,Rectangle,Diamond,Triangle,Hexagon,Star,Limping,Prism,Implied,Solid,Alternating,Concentric,Radiating,NoShape,Square,Rectangle,Diamond,Limping,Custom,Square,Rectangle,Diamond,Prism,Triangle,Hexagon,Star,Limping,T31212,T3464,T488,T33434,T33336,T33344,T3636,T4612,T333333_33434,NoShape,Square,Rectangle,Diamond,Triangle,Hexagon,Star,Limping,Prism,Star,TrumpSuit,Rank,Suit,TrumpValue,TrumpRank,Pieces,Pips,Rows,Columns,Turns,Moves,Trials,MovesThisTurn,Phases,Vertices,Edges,Cells,Players,Active,Sites,Adjacent,Neighbours,Orthogonal,Diagonal,Off,Groups,Liberties,Steps,To,From,Group,Stack,Territory,Move,EndSite,Piece,Player,Pending,Around,Crossing,Direction,Distance,Axial,Horizontal,Vertical,Angled,Slash,Slosh,Group,Incident,Row,Column,Phase,Cell,Edge,State,Empty,LineOfSight,Occupied,Hand,Start,Track,Winning,Visible,Masked,Invisible,Side,Board,Top,Bottom,Left,Right,Inner,Outer,Corners,ConcaveCorners,ConvexCorners,Major,Minor,Centre,Hint,ToClear,LineOfPlay,Pending,Playable,LastTo,LastFrom,Track,Bet,FromTo,Hop,Leap,Propose,Vote,Promote,Remove,Select,Set,Shoot,Pass,PlayCard,Add,Claim,Slide,Step,Direction,NextPlayer,Pending,Value,Score,Visible,Masked,Invisible,Count,State,TrumpSuit,Counter,Var,Players,Pieces,Control,Domino,Distance,Moves,Captures,Die,Direction,Piece,Player,Site,Random,Stack,AllInvisible,Team,Amount,Score,Count,Cost,Phase,Deck,NoBasis,Triangular,Square,Hexagonal,T33336,T33344,T33434,T3464,T3636,T4612,T488,T31212,T333333_33434,SquarePyramidal,HexagonalPyramidal,Wheel,Circle,Spiral,Dual,Brick,Mesh,Morris,Celtic,QuadHex,CentreSite,LeftSite,RightSite,Topsite,BottomSite,FirstSite,LastSite,Cell,Edge,Vertex,Hint,Empty,NotEmpty,Own,NotOwn,Enemy,NotEnemy,AllPlayers,Rows,Columns,AllDirections,HintRegions,Layers,Diagonals,SubGrids,Regions,Vertices,Corners,Sides,SidesNoCorners,AllSites,Touching,Orthogonal,Diagonal,Off,Adjacent,All,NoShape,Custom,Square,Rectangle,Triangle,Hexagon,Cross,Diamond,Prism,Quadrilateral,Rhombus,Wheel,Circle,Spiral,Wedge,Star,Limping,Polygon,Vertex,Edge,Cell,F,B,L,R,Square,Triangular,Hexagonal,Track,Joker,Ace,Two,Three,Four,Five,Six,Seven,Eight,Nine,Ten,Jack,Queen,King,Dominoes,Cards,Clubs,Spades,Diamonds,Hearts,Alternating,Simultaneous,InTurn,InGame,Positional,Situational,Infinite,Win,Loss,Draw,Tie,Abandon,Crash,Neutral,P1,P2,P3,P4,P5,P6,P7,P8,P9,P10,P11,P12,P13,P14,P15,P16,Team1,Team2,Team3,Team4,Team5,Team6,Team7,Team8,Team9,Team10,Team11,Team12,Team13,Team14,Team15,Team16,Each,Shared,All,Any,Mover,Next,Prev,NonMover,Enemy,Ally,NonAlly,Partner,NonPartner,NonNeutral,Player,StartOfMove,EndOfMove,StartOfTurn,EndOfTurn,StartOfRound,EndOfRound,StartOfPhase,EndOfPhase,StartOfGame,EndOfGame,StartOfMatch,EndOfMatch,StartOfSession,EndOfSession,Sites,Moves,All,Angled,Adjacent,Axial,Orthogonal,Diagonal,Off,SameLayer,Upward,Downward,Rotational,N,E,S,W,NE,SE,NW,SW,NNW,WNW,WSW,SSW,SSE,ESE,ENE,NNE,CW,CCW,In,Out,U,UN,UNE,UE,USE,US,USW,UW,UNW,D,DN,DNE,DE,DSE,DS,DSW,DW,DNW,N,NNE,NE,ENE,E,ESE,SE,SSE,S,SSW,SW,WSW,W,WNW,NW,NNW,N,NNE,NE,E,SSE,SE,S,SSW,SW,W,NW,NNW,WNW,ENE,ESE,WSW,CW,Out,CCW,In,UNW,UNE,USE,USW,DNW,DNE,DSE,DSW,U,UN,UW,UE,US,D,DN,DW,DE,DS,Forward,Backward,Rightward,Leftward,Forwards,Backwards,Rightwards,Leftwards,FL,FLL,FLLL,BL,BLL,BLLL,FR,FRR,FRRR,BR,BRR,BRRR,SameDirection,OppositeDirection,Out,CW,In,CCW,D,DN,DNE,DE,DSE,DS,DSW,DW,DNW,U,UN,UNE,UE,USE,US,USW,UW,UNW,Checkered,Colour,Shape,StyleThickness,Style,Board,Animation,HandScale,Curves,MaskedColour,ColourFromState,Colour,Families,Background,Foreground,Rename,ExtendName,AddStateToName,Reflect,ByValue,Scale,Style,Colour,Colour,Pits,PlayerHoles,RegionOwner,Cost,Check,State,Value,Piece,Edges,Score,Shape,AsHoles,Sites,Cell,Symbol,InnerEdges,OuterEdges,Phase0,Phase1,Phase2,Phase3,Symbols,Vertices,White,Black,Grey,LightGrey,VeryLightGrey,DarkGrey,VeryDarkGrey,Dark,Red,Green,Blue,Yellow,Pink,Cyan,Brown,DarkBrown,Purple,Turquoise,Orange,DarkOrange,LightRed,DarkRed,LightGreen,DarkGreen,LightBlue,VeryLightBlue,DarkBlue,IceBlue,Gold,Silver,Bronze,GunMetal,HumanLight,HumanDark,Cream,DeepPurple,PinkFloyd,BlackSabbath,KingCrimson,MoodyBlues,TangerineDream,Piece,Tile,Card,Die,Domino,LargePiece,ExtendedShogi,Board,Hand,Deck,Dice,Boardless,ConnectiveGoal,Mancala,PenAndPaper,Pyramidal,Spiral,Isometric,Puzzle,GraphPuzzle,LineSegment,PuzzlePenAndPaper,RegionPuzzle,Agon,Backgammon,Chess,ChineseCheckers,Connect4,Goose,Go,Graph,HoundsAndJackals,Janggi,Lasca,Pachisi,Ploy,Scripta,Shogi,SnakesAndLadders,Surakarta,Tafl,Xiangqi,UltimateTicTacToe,Futoshiki,Hashi,Kakuro,Sudoku,BasicController,PyramidalController,All,Inner,Outer,Interlayer,Thin,Thick,ThinDotted,ThickDotted,ThinDashed,ThickDashed,Hidden,Default,Ground,Reverse,Fan,None,Backgammon,Ring,None,Corner,Middle,Always,Never,AtEnd},
  ndkeywordstyle=\color{dviolet}\bfseries,
  identifierstyle=\color{black},
  sensitive=true,   % need case-sensitivity for different keywords
  comment=[l]{//},
  commentstyle=\color{dred}\ttfamily,
  stringstyle=\color{dgreen}\ttfamily,
  morestring=[b]',
  morestring=[b]",
  escapechar=@,
  showstringspaces=false,
  xleftmargin=1pt,xrightmargin=1pt,
  breaklines=true,basicstyle=\ttfamily\small,backgroundcolor=\color{colorex},inputencoding=utf8/latin9,texcl
}

\newcommand{\core}{ 
  \medskip \begin{tcolorbox}[
    enhanced,breakable,
    boxsep=0pt,top=-4pt,bottom=-4pt,left=1mm,right=1mm,
    toprule=0.1mm,leftrule=0.1mm,rightrule=0.25mm,bottomrule=0.25mm,shadow={0.2mm}{-0.2mm}{0mm}{dgray},
    overlay unbroken and first={\node (logo) at ([xshift=6mm,yshift=-5mm]frame.north west) {}; %\draw[black,line width=1.5pt] (logo) -- ([xshift=4mm,yshift=1.5mm]frame.south west);  
    },
    colframe=dgray,titlerule=-0.2mm,toptitle=3mm,coltitle=black,fonttitle=\bfseries,
    lines before break=6, pad at break*=2pt
}
\newcounter{cntEx}
\newcounter{cntNaturalLanguage}

\ifx\bw\undefined
  \lstnewenvironment{ludii}[1][]{\lstset{language=ludii,#1}}{}
  \lstnewenvironment{syntax}{\lstset{}}{}
  \newenvironment{boxex}
    {\stepcounter{cntEx} \core ,colback=colorex,title style={color=colorex} %\thecntEx
    ]}
    {\end{tcolorbox}} %{\vspace{-0.1cm}\end{tcolorbox} ~ \vspace{-0.2cm}}
\definecolor{v2lgray}{gray}{0.85}
\definecolor{dgray}{rgb}{0.4,0.4,0.4}
\definecolor{dblue}{RGB}{0,0,99}
\definecolor{dred}{RGB}{150,6,54}
\definecolor{dgreen}{RGB}{47,135,7}
\definecolor{dviolet}{RGB}{102,0,153}
\definecolor{mblue}{RGB}{0,0,180}
\definecolor{colorex}{HTML}{F0F8FF}  %{FFE3BE} DFEFFF F0F8FF
\definecolor{grey1}{rgb}{0.9,0.9,0.9}

\begin{document}
\title{Exploring RL-based LLM Training for Formal Language Tasks with Programmed Rewards}
\titlerunning{Exploring RL-based LLM Training for Formal Language Tasks}
% \institute{Department of Advanced Computing Sciences, Maastricht University,\\ Paul-Henri Spaaklaan 1, 6229 EN Maastricht, the Netherlands

% Option 1
\author{Alexander G. Padula\inst{1,2} \and Dennis J.N.J. Soemers\inst{2}}
\authorrunning{A.G. Padula and D.J.N.J. Soemers}
\institute{Department of Computer Science, ETH Zurich \and Department of Advanced Computing Sciences, Maastricht University
\email{apadula@student.ethz.ch}, \email{dennis.soemers@maastrichtuniversity.nl}}

% Option 2
% \author[1]{Alexander G. Padula}
% \author[2]{Dennis J.N.J. Soemers}

% \affil[1]{Department of Computer Science, ETH Zurich, \\
% Universitätstrasse 6, 8092 Zurich, Switzerland \\
% \email{apadula@student.ethz.ch}}
% \affil[2]{Department of Advanced Computing Sciences, Maastricht University, \\
% Paul-Henri Spaaklaan 1, 6229 EN Maastricht, the Netherlands \\
% \email{dennis.soemers@maastrichtuniversity.nl}}

%
\maketitle              % typeset the header of the contribution
\begin{abstract}
Proximal Policy Optimization (PPO) is commonly used in Reinforcement Learning from Human Feedback to align large language models (LLMs) with downstream tasks. This paper investigates the feasibility of using PPO for direct reinforcement learning (RL) from explicitly programmed reward signals, as opposed to indirect learning from human feedback via an intermediary reward model. We focus on tasks expressed through formal languages, such as mathematics and programming, where explicit reward functions can be programmed to automatically assess the quality of generated outputs. We apply this approach to a sentiment alignment task, a simple arithmetic task, and a more complex game synthesis task. The sentiment alignment task replicates prior research and serves to validate our experimental setup. Our results show that pure RL-based training for the two formal language tasks is challenging, with success being limited even for the simple arithmetic task. We propose a novel batch-entropy regularization term to aid exploration, although training is not yet entirely stable. Our findings suggest that direct RL training of LLMs may be more suitable for relatively minor changes, such as alignment, than for learning new tasks altogether, even if an informative reward signal can be expressed programmatically.
%The abstract should briefly summarize the contents of the paper in 150--250 words.

\end{abstract}
\section{Introduction}
Program synthesis models such as Copilot \cite{chen2021evaluating_copilot} have quickly become indispensable by improving productivity and making complex tasks more accessible. 
Significant advancements in this field have been achieved by training general-purpose Large Language Models (LLMs) to reproduce source code from widely used programming languages. 
Current state-of-the-art coding models \cite{li2023starcoder,luo2023wizardcoder,rozière2024llamacode} are trained using an auto-regressive next-token prediction objective, which maximizes the probability of predicting the next token in a sequence. Despite their success across a wide range of language processing tasks, next-token prediction suffers from the inherent limitation of being a surrogate objective, which can at times diverge from a task's true goals. 
%When multiple valid solutions exist for a task, such as implementing a function, next-token prediction penalizes the model for deviating from solutions that are overrepresented in the training data, even if the model could have otherwise generated an alternative that is equivalent or superior. 
When multiple valid solutions exist for a task, such as implementing a function, next-token prediction penalizes the model for deviating from solutions that are overrepresented in the training data, even if other solutions may be equivalent or superior. 
Moreover, coding models trained with a next-token objective are not grounded in the outcomes of executing the code they generate. This disconnect can exacerbate existing tendencies to produce near misses, where the generated code superficially resembles a correct solution but contains subtle errors that prevent successful execution \cite{chen2021evaluating_copilot,austin2021programsynthesislargelanguage}.
% arguably limit the potential for controllability and alignment with downstream tasks \cite{bubeck2023sparks,lecun2023,bachmann2024pitfalls}. Unlike intelligent beings, such as animals or humans, traditionally trained LLMs do not learn from countless interactions with environments; they are instead confined to inferring statistical patterns in language. This limitation opens them to criticisms that they reproduce linguistic form whilst failing to grasp function, ultimately rendering outputs that mimic reasoning without deeper understanding \cite{marcus2018rebooting,chomsky2019limits}.

Reinforcement Learning (RL) \cite{Sutton_2018_RL} emerges as a natural paradigm to ground models by training directly on a task's true goals. However, in many (natural) language tasks it is often difficult to explicitly program a reliable reward signal to quantify useful properties (e.g., a response's helpfulness or accuracy).
%However, defining a useful reward function in language processing domains is often challenging. For digital assistants and natural language tasks, it is difficult to explicitly program a signal to quantify helpfulness or accuracy. 
RL from Human Feedback (RLHF) circumvents this challenge by training an intermediary critic model on a limited number of human evaluations collected from users or reviewers \cite{ziegler2020finetuning}. 
%The trained critic model evaluates the base model's responses, aligning this process closer to a classical Reinforcement Learning setting where a policy gradient method, typically a form of Proximal Policy Optimization (PPO), is used to update the base model’s weights in order to maximize the reward obtained from the reward model. 
The trained critic model produces rewards as evaluations of the base model's outputs, enabling the use of RL to further train the model so as to generate outputs that maximize the intermediate rewards.
Despite its complexity, RLHF has become a popular and effective method for aligning pre-trained LLMs with downstream tasks \cite{openai2021aligning,openai2022chatgpt,Llama3_2024}.
%Its widespread use has led to inclusion in the open-source Hugging Face ecosystem as part of the Transformers Reinforcement Learning (TRL) library \cite{vonwerra2022trl}.

In the domain of programming, and more broadly structured languages, however, there is a distinctive opportunity to employ a more direct and contextually appropriate training objective. Unlike natural language, code can be executed, and its effects can be automatically evaluated and compared, offering a pathway to explicitly programmed domain-specific reward functions. This eliminates the need for human-in-the-loop evaluations and reward models, aligning the training process more closely with how humans experiment and learn coding through a continuous process of trial and error. This paper explores the idea of exploiting this unique aspect of structured languages to train LLMs using reward signals obtained from explicitly programmed functions as a direct training objective.

Recent research on applying RL to LLMs \cite{le2022coderl,shojaee2023execution,uesato2022solving} has often relied on custom implementations. In contrast, this paper makes only minimal adjustments to the existing RLHF implementation in Hugging Face's Transformers Reinforcement Learning (TRL) library \cite{vonwerra2022trl}. This approach aims to simplify reproducibility by utilizing an established deep learning ecosystem.\footnote{TRL fork with (batch-)entropy regularization: \url{https://github.com/PadLex/trl}.}\footnote{Source code for experiments can be found at: \url{https://github.com/PadLex/Reinforcement-Learning-from-Explicitly-Programmed-Reward-Signals/tree/main}.}

\section{Background}

Auto-regressive text generation tasks, such as program synthesis, can be modeled, following the standard RL problem formulation, as a finite-horizon Markov Decision Process (MDP). 
At any time step $t$, a state $s_t \in \mathcal{S}$ from a state space $\mathcal{S}$ is characterized by the sequence of non-masked tokens from the beginning of the sequence up to \( t \). This representation captures the necessary context for subsequent token generation. The action $a_t \in \mathcal{A}$ from an action space $\mathcal{A}$ at time $t$ corresponds to selecting the next token to add from a predefined vocabulary, extending the current state \( s_t \) by one token. A policy $\pi_{\theta}$, parameterized by tunable parameters \( \theta \), outputs a probability distribution over \( \mathcal{A} \) conditioned on the current state \( s_t \). During training, the parameters \( \theta \) are adjusted to maximize observed rewards. A reward function \( R \) provides a scalar signal $R(\tau)$ for a trajectory \( \tau = (s_1, a_1, s_2, a_2, \ldots, s_n, a_n) \). Given the auto-regressive nature of the task, where the final state \( s_n \) encapsulates the entire sequence, the reward can also be expressed solely in terms of \( s_n \) as \( R(s_n) \). In auto-regressive text generation, a discount factor of 1 (i.e., no discounting) is typically used as the MDP is finite and there is no explicit preference for shorter solutions.
% \begin{itemize}
%     \item \textbf{State} \( s_t \in \mathcal{S} \): A state \( s_t \) at any time step $t$ is characterized by the sequence of non-masked tokens from the beginning of the sequence up to \( t \). This representation captures the necessary context for subsequent token generation.
%     \item \textbf{Action} \( a_t \in \mathcal{A} \): The action at time step \(t\) involves selecting the next token to add to the sequence from a predefined vocabulary. This action extends the current state \( s_t \) by one token.
%     \item \textbf{Policy} \( \pi_\theta(a_t \mid s_t) \): This stochastic policy, parameterized by tunable parameters \( \theta \), outputs a probability distribution over the action space \( \mathcal{A} \) conditioned on the current state \( s_t \). During training, the parameters \( \theta \) are adjusted to maximize observed rewards.
%     \item \textbf{Reward} \( R(\tau) \) or \( R(s_n) \): The reward function \( R \) provides a scalar signal for a trajectory \( \tau = (s_1, a_1, s_2, a_2, \ldots, s_n, a_n) \). Given the auto-regressive nature of the task, where the final state \( s_n \) encapsulates the entire sequence, the reward can also be expressed solely in terms of \( s_n \) as \( R(s_n) \). In auto-regressive text generation, a discount factor of 1 (i.e., no discounting) is typically used since the MDP is finite and there is usually no preference for shorter solutions.
% \end{itemize}

The field of RL \cite{Sutton_2018_RL} develops algorithms that update a policy's parameters based on experience, so as to maximize the rewards collected by the policy in future trajectories. Policy gradient methods define a differentiable objective function \( L(\theta) \), which can be maximized using optimizers such as Adam or Stochastic Gradient Ascent to improve the policy's performance. Proximal Policy Optimization (PPO) \cite{schulman2017proximal} has become the de facto standard for RL-based training of LLMs following its use in RLHF. 
%Given its popularity, PPO is a natural choice for a baseline policy optimization method in this paper.
Schulman et al. \cite{schulman2017proximal} originally proposed two variants of PPO, both aiming to maximize the surrogate objective \(L^{CPI}(\theta)\):
$$
r_t(\theta) = \frac{\pi_\theta(a_t | s_t)}{\pi_{\theta_\text{old}}(a_t | s_t)},\qquad L^{CPI}(\theta) = \hat{\mathop{\mathbb{E}}}_t[r_t(\theta) \hat{A}_t],
$$
where the expectation estimator \(\hat{\mathop{\mathbb{E}}}_t[\dots]\) measures the empirical average over a finite batch of samples. The ratio \(r_t(\theta)\) moderates the extent of the policy updates based on how much more or less likely the action \(a_t\) is under the new policy \(\pi_\theta\) compared to the previous policy \(\pi_{\theta_\text{old}}\). Compared to a traditional policy gradient objective \cite{SpinningUp2018}, this surrogate objective gives more conservative updates and generally leads to a more stable learning process \cite{schulman2017proximal}. The advantage \( \hat{A}_t \) estimates the relative benefit of taking the action \(a_t\) compared to all other actions available in \(s_t\), weighted by their probability under \(\pi_\theta\). It is used to isolate the effect of specific actions from the general quality of the states in which they are taken. Using the Bellman equation \cite{BellmanDynamicProgramming,Sutton_2018_RL}, we can express the advantage as \( \hat{A}_t = R_t + V(s_{t+1}) - V(s_t) \), where $V(s)$ is the value (expected sum of future rewards) of a state $s$. A value function (e.g., neural network) trained to minimize the mean squared error between predicted and observed values estimates $V$.
%where:
% \begin{itemize}
%     \item The expectation estimator \(\hat{\mathop{\mathbb{E}}}_t[\dots]\) measures the empirical average over a finite batch of samples.
%     \item The probability ratio \(r_t(\theta)\) moderates the extent of the policy updates based on how much more or less likely the action \(a_t\) is under the new policy \(\pi_\theta\) compared to the previous policy \(\pi_{\theta_\text{old}}\). Compared to a traditional policy gradient objective \cite{SpinningUp2018}, this surrogate objective is more conservative in its updates and generally leads to a more stable learning process \cite{schulman2017proximal}.
%     \item The advantage \( \hat{A}_t \) estimates the relative benefit of taking the action \(a_t\) compared to all other actions available in \(s_t\), weighted by the probability of selecting them under \(\pi_\theta\). It is used to isolate the effect of specific actions from the general quality of the states in which they are taken. Using the Bellman equation \cite{BellmanDynamicProgramming,Sutton_2018_RL}, we can express the advantage as \( \hat{A}_t = R_t + V(s_{t+1}) - V(s_t) \), where $V(s)$ is the value (expected sum of future rewards) of being in a state $s$. In practice, \( V \) is estimated by a value function (e.g., neural network) trained to minimize the mean squared error between predicted and observed values.
% \end{itemize}

While both variants of PPO aim to maximize \(L^{CPI}(\theta)\), they differ in the constraints that they employ to avoid overly aggressive gradient updates. The \textit{Clipped Surrogate Objective} variant of PPO disincentivizes \(r_t(\theta)\) from moving outside of the interval \([1-\epsilon, 1+\epsilon]\):

\[
L^{CLIP}(\theta) = \hat{\mathop{\mathbb{E}}}_t \left[ \min\left(r_t(\theta) \hat{A}_t, \text{clip}(r_t(\theta), 1-\epsilon, 1+\epsilon) \hat{A}_t\right) \right]
\]
The \textit{Adaptive KL Penalty Coefficient} variant of PPO penalizes large changes to the policy \( \pi_\theta(a_t \mid s_t) \) by measuring the Kullback–Leibler (KL) divergence \cite{KullbackLeibler1951} between \( \pi_{\theta_\text{old}}(a_t \mid s_t) \) and \( \pi_{\theta}(a_t \mid s_t) \): %The KL penalty coefficient \(\beta_\text{KL}\) can change during training to adapt the weight of the penalty:
\[
L^{KLP}(\theta) = \hat{\mathop{\mathbb{E}}}_t \left[ r_t(\theta) \hat{A}_t - \beta_\text{KL} KL[\pi_{\theta_\text{old}}(a_t \mid s_t), \pi_{\theta}(a_t \mid s_t)] \right]
\]
An \textit{entropy regularization} term \(L^{ENT}\) is often added to the objective function to encourage exploration and smoothen the optimization landscape \cite{Ahmed_2019_Understanding}:
\begin{equation} \label{Eq:EntropyRegularization}
L^{ENT}(\theta) = - \beta_\text{ENT} \hat{\mathop{\mathbb{E}}}_t \left[ \sum_{a \in \mathcal{A}} \pi_\theta(a \mid s_t) \log{\pi_\theta}(a \mid s_t) \right]
\end{equation}
%This paper takes the TRL implementation \cite{vonwerra2022trl} of PPO as baseline. 
%TRL implements a variant of PPO designed by Ziegler et al. (2020) \cite{ziegler2020finetuning} for RLHF, which differs from common PPO implementations by using both a Clipped Surrogate Objective and an Adaptive KL Penalty. 
%Furthermore, it drops the entropy regularization term and uses a KL divergence relative to the initial policy, instead of the policy at the previous training step. 
% We take the TRL implementation \cite{vonwerra2022trl} of PPO as baseline, which implements a varient of PPO standard for RLHF \cite{ziegler2020finetuning}.

We use the TRL implementation \cite{vonwerra2022trl} of PPO as our baseline, which is based on a variant of PPO that has become the standard for RLHF \cite{ziegler2020finetuning}. It differs from common (non-RLHF) implementations by using both a Clipped Surrogate Objective and an Adaptive KL Penalty, and dropping the entropy regularization term. Furthermore, its KL divergence is relative to the initial policy, instead of the policy at the previous training step. 
This is because in RLHF, the initial policy (usually a foundation model pre-trained on a substantial amount of data) is assumed to be close to the final one, and care must be taken to ensure that the policy does not fool the reward model by generating out-of-distribution samples.

\section{Batch-Entropy Regularization}

The standard entropy regularization term of \refequation{Eq:EntropyRegularization} encourages exploration by penalizing the policy for drifting away from the uniform policy on the basis of individual states: within a training batch, every state with a highly non-uniform policy receives a sizeable penalty. However, a potential failure mode of RL-based training may not just be a lack of exploration on a per-state basis, but rather a lack of exploration across the state space. A strong policy will often need to put most of its probability mass on a single (best) action per state---a solution that conflicts with the standard entropy regularization---but would typically be expected to still pick different actions across different states.

We propose a novel \textit{batch-entropy regularization} term, designed to encourage the policy to choose diverse actions for different states within the same batch, without penalizing it for having a highly non-uniform policy for individual states: 
%Unlike the conventional entropy bonus, which promotes the exploration of different actions by encouraging stochasticity, the batch-entropy bonus allows the policy to potentially achieve full convergence to a deterministic state-action mapping, while still fostering the exploration of diverse actions between states.
\begin{equation} \label{Eq:BatchEntropyRegularization}
    L^{BENT}(\theta) = - \beta_\text{BENT} \sum_{a \in \mathcal{A}} \hat{\mathbb{E}}_{s_t \sim B} \left[ \pi(a \mid s_t) \right] \log \hat{\mathbb{E}}_{s_t \sim B} \left[ \pi(a \mid s_t) \right],
\end{equation}
where $\hat{\mathbb{E}}_{s_t \sim B}$ denotes the empirical mean over all states $s_t$ in a batch $B$. Such a batch-entropy term was previously used to analyze and evaluate the behavior of trained RL models \cite{Dereventsov_2024_Examining}, but our use as a regularization term is novel.

\section{Language Model Fine-tuning Tasks}

In this paper, we consider three different RL-based fine-tuning tasks for language models. Firstly, a sentiment alignment task (\refsubsection{Subsec:SentimentAlignmentTask})---for which successful results are known to be feasible from prior work \cite{HuggingFace_2022_sentiment}---is used to verify the correctness and compatibility of our TRL-based implementation and experiment setup. Secondly, a synthetic arithmetic task (\refsubsection{Subsec:SyntheticArithmeticTask}) is used as a formal language task which is simple enough to rapidly generate substantial training data. Thirdly, we consider the complex task of synthesising (novel) games in Ludii's formal game description language \cite{Browne_2020_LLR,Piette_2020_Ludii} (\refsubsection{Subsec:LudiiGameSynthesisTask}). This task is particularly compelling due to the scarcity of training samples---which limits the effectiveness of supervised learning---and the availability of established reward metrics to assess the quality of newly generated Ludii games \cite{Todd_2024_GAVEL}.

\subsection{Sentiment Alignment Task} \label{Subsec:SentimentAlignmentTask}

We initially seek to replicate results from prior research \cite{HuggingFace_2022_sentiment} in order to verify that the models we employ are compatible with TRL and confirm that the modifications we make to TRL--—adding entropy and batch-entropy losses, removing the KL penalty, and replacing the trained reward model with a programmed reward function—--do not compromise the integrity of the experimental setup.

The task is to fine-tune GPT-2 \cite{radford2019language_gpt2} to generate positive movie reviews using RL. Initially, GPT-2 is pre-trained using a conventional masked language modeling (MLM) objective on the Stanford IMDB dataset \cite{StanfordIMDB}. Then, using PPO, the model is trained to complete reviews from the dataset while imbuing them with a positive sentiment. As part of the RLHF process, generated samples are evaluated by a reward model. For this purpose, the research we are reproducing employs a variant of DistilBERT \cite{sanh2020distilbert} that was fine-tuned on user-labeled reviews in the IMDB dataset. As an alternative training method, we also replace the conventional reward model with an automated signal. For this, we use the VADER \cite{Hutto_2014_VADER} implementation from NLTK \cite{BirdKleinLoper09_NLTK}, a rule-based sentiment analysis algorithm that returns a score between -1 and 1, which we use as a reward signal that quantifies how negative or positive the generated reviews are.

\subsection{Synthetic Arithmetic Task} \label{Subsec:SyntheticArithmeticTask}

We define a simple arithmetic task designed to elucidate the potential advantages offered by RL-based training over a traditional MLM objective. In this task, \( n = 5 \) coefficients, \( c_1, c_2, \ldots, c_n \), are independently and uniformly drawn from the set of integers \( \{0, 1, \ldots, 9\} \). These coefficients are summed to form an initial expression, \( Y_0 = c_1 + c_2 + \ldots + c_n \). The task simplifies \( Y_0 \) through a series of \( n \) steps, where at each step \( i \), two randomly chosen, non-simplified terms from \( Y_{i-1} \) are resolved (i.e., added together), resulting in \( Y_i \). This process is repeated until the final expression, \( Y_n \), is a single integer: the sum of all original coefficients.

Importantly, due to the random order of summations, the sequence of intermediate expressions \( Y_1, Y_2, \ldots, Y_n \) is non-deterministic. This randomness restricts the effectiveness of an imitation learning strategy in minimizing its loss, as it cannot leverage consistent sequential dependencies typically exploited in MLM tasks. For example, the sum \( 6 + 10 + 7 + 1 + 3 \) might first simplify to \( 6 + 17 + 1 + 3 \), then to \( 23 + 1 + 3 \), followed by \( 23 + 4 \), and finally to \( 27 \), with each step involving the addition of randomly selected terms from the previous expression. This task has properties that mirror program synthesis, where multiple equally valid solutions exist and their correctness can be quantified.

The reward function for RL-based training is designed to quantify the accuracy of the model's generated expressions relative to the target expressions. It assigns a scalar reward based on the absolute difference between the summed value of the generated expression \( G_i \) and the correct expression \( Y_i \):
\[
R(G_i, Y_i) = \frac{2}{1 + \exp\left(\frac{\lvert G_i - Y_i \rvert}{10}\right)}
\]
% \[
% R(G_i, Y_i) = 2 \times \left( 1 + \exp\left(\frac{\lvert G_i - Y_i \rvert}{10}\right) \right)^{-1}
% \]
The reward is $0$ in cases where \( G_i \) is an invalid expression. This function ensures that smaller errors lead to higher rewards, and significant errors, particularly from invalid expressions, result in low rewards. The offset sigmoid function ensures that the reward scales smoothly between 0 and 1, providing a non-flat reward landscape even early on in training. Note how teacher forcing prevents the accumulation of errors between, but not within, expressions. In other words, \( G_{i-1} \) is discarded and the model is instead shown \( Y_{i-1} \) when computing \( G_i \). So if the model makes a mistake when generating \( G_{i-1} \), that mistake will not carry over when it prompted to generate \( G_i \). In contrast to MLM objectives, this reward signal is invariant to changes in the order in which terms are summed.

\subsection{Ludii Game Synthesis Task} \label{Subsec:LudiiGameSynthesisTask}
Ludii \cite{Piette_2020_Ludii} is a general game playing system with a domain-specific language (DSL) for describing rules of games \cite{Browne_2020_LLR}. Any description of rules in this language can be compiled into a runnable game by the system. This DSL describes games as trees of \textit{ludemes}, which are high-level keywords corresponding to common board game concepts such as \textit{board}, \textit{is empty}, \textit{is line}, \textit{step}, \textit{slide}, and so on. For an example, see the game description for the connection game \textit{Hex}:

\begin{boxex}
\begin{ludii}
(game "Hex"  
  (players 2)  
  (equipment { 
      (board (hex Diamond 11)) 
      (piece "Marker" Each) 
      (regions P1 {(sites Side NE) (sites Side SW)}) 
      (regions P2 {(sites Side NW) (sites Side SE)})
  })  
  (rules 
    (play (move Add (to (sites Empty))))
    (end (if (is Connected Mover) (result Mover Win)))
  )
)
\end{ludii} 
\end{boxex}

Generating games in this DSL is ideally suited to exploring how a direct RL process can overcome limitations arising from limited data availability. Although board game representations in this DSL are succinct enough to fit within the context length of modern LLMs and can be directly compiled into fully playable and testable games, there are only in the order of 1000 existing board games implemented in the Ludii DSL. The scarcity of available data makes it challenging to train LLMs to learn Ludii using traditional supervised training methods.

To ease the model into learning the Ludii DSL, we define a fill-in-the-middle task. In this task, uniformly randomly sampled parentheticals are removed from game descriptions, and the model is trained to generate the missing sections. In this way, the dataset will range from simple prompts requiring the model to only fill in a small portion of a game, all the way to requiring the model to complete a whole game from scratch when the root parenthetical is sampled. The following example shows pre- and suffixes for the \textit{Hex} game description, with the final part of the \texttt{equipment} section of the description having been removed:

% \begin{figure}[!ht]
%     \centering
%     \includegraphics[width=\linewidth]{images/hex.png}
%     \caption{An example representation of the game of Hex in the Ludii DSL}
%     \label{fig:hex}
% \end{figure}

% \begin{figure}[!ht]
%     \centering
%     \includegraphics[width=\linewidth]{images/fill_hex.png}
%     \caption{A prompt from the fill-in-the-middle Ludii task that was sampled from the Hex game description}
%     \label{fig:hex}
% \end{figure}

\begin{boxex}
\begin{ludii}
### PRE:
(game "Hex"  
  (players 2)  
  (equipment { 
      (board (hex Diamond 11)) 
      (piece "Marker" Each) 
      (regions P1 {(sites Side NE) (sites Side SW)}) 

### SUF:
  })  
  (rules 
    (play (move Add (to (sites Empty))))
    (end (if (is Connected Mover) (result Mover Win)))
  )
)
\end{ludii} 
\end{boxex}

While the quality of a game description is more challenging to objectively quantify than the correctness of, e.g., a simple program or a solution for the arithmetic task, it is still possible to program a reasonable reward function. Inspired by fitness functions used by prior work on evolutionary game generation \cite{Browne_2009_PhD,Todd_2024_GAVEL}, we use a reward function based on the following five criteria:
\begin{enumerate}
    \item \textbf{Compilability} \(C: S \mapsto \{0, 1\}\): A binary signal indicating whether the game compiles, i.e., whether the game is syntactically valid and avoids semantic errors such as using a piece that was not defined in the \texttt{equipment} ludeme.
    \item \textbf{Playability} \(P: S \mapsto \{0, 1\}\):  A binary signal indicating whether or not moves can be made without crashing.
    \item \textbf{Balance} \(B: S \mapsto [0, 1]\): A continuous signal defined as the largest difference in winrates between any pair of players. For example, it returns 1 if all players won the same number of games, and 0 if one player won them all.
    \item \textbf{Completion Rate} \(F: S \mapsto [0, 1]\): The fraction of games that terminated within 500 turns.
    \item \textbf{Decisiveness} \(D: S \mapsto [0, 1]\): The fraction of games that did not end in a draw. It returns 1 if all the games ended with a winner or loser.
\end{enumerate}
The first criterion can be evaluated simply by having Ludii try to compile any given game description, whereas the other four require playing the game. We use 100 playthroughs (per generated game description) in which moves are selected uniformly at random to compute these criteria. Using games played between stronger agents could lead to more informative signals, but would have been prohibitive in terms of computation time. Ultimately, for any generated game description $s$, we use a reward of $R(s) = 0$ if $s$ cannot be compiled (i.e., if $C(s) = 0$), $R(s) = 0.1$ if it is not playable (i.e., if $P(s) = 0$), or the geometric mean $\frac{1}{3} \left( B(s)^{\frac{1}{3}} + F(s)^{\frac{1}{3}} + D(s)^{\frac{1}{3}}\right)$ of the remaining three criteria otherwise.
% \[
% R(s) = 
% \begin{cases} 
% 0 & \text{if } C(s) = 0 \\
% 0.1 & \text{else if } P(s) = 0 \\
% \frac{1}{3} \left( B(s)^{\frac{1}{3}} + F(s)^{\frac{1}{3}} + D(s)^{\frac{1}{3}} \right) & \text{otherwise}
% \end{cases}
% \]

\section{Experiments}

% \begin{figure}[t]
%     \centering
%     \includegraphics[width=.7\linewidth]{images/imdb_model_comparison_BERT_silver-smoke-184_divine-rain-209_hardy-darkness-185.png}
%     \caption{Comparison between three different base models, being trained with PPO for the sentiment alignment task, using DistillBERT as a trained reward model.}
%     \label{fig:imdb_model_comparison_bert}
% \end{figure}

% \begin{figure}[t]
%     \centering
%     \includegraphics[width=.7\linewidth]{images/imdb_model_comparison_VADER.png}
%     \caption{Comparison between three different base models, being trained with PPO for the sentiment alignment task, using VADER as programmatic reward function.}
%     \label{fig:imdb_model_comparison_vader}
% \end{figure}

\begin{figure}[t]
    \centering
    \includegraphics[width=.7\linewidth]{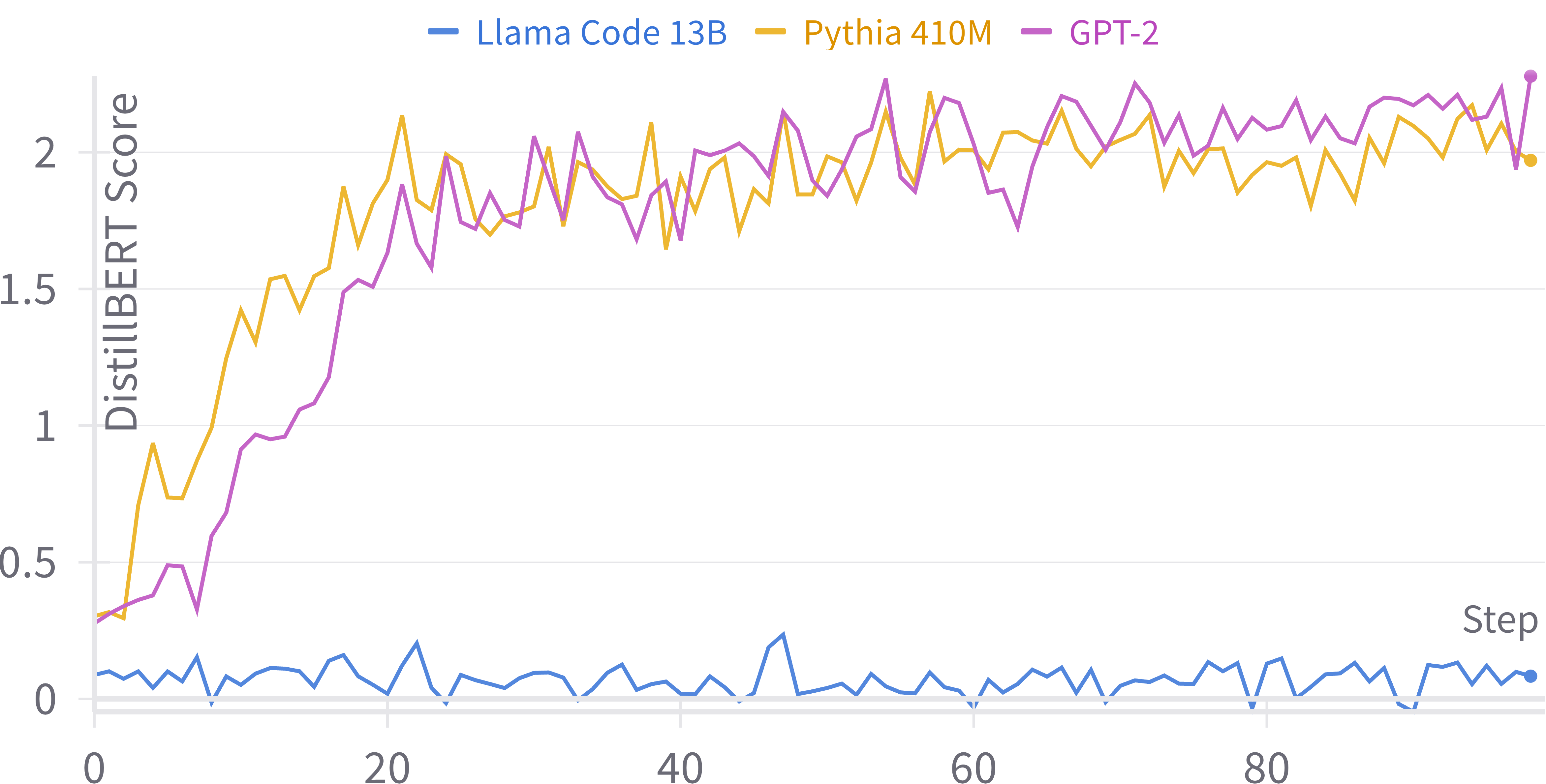}
    \includegraphics[width=.72\linewidth]{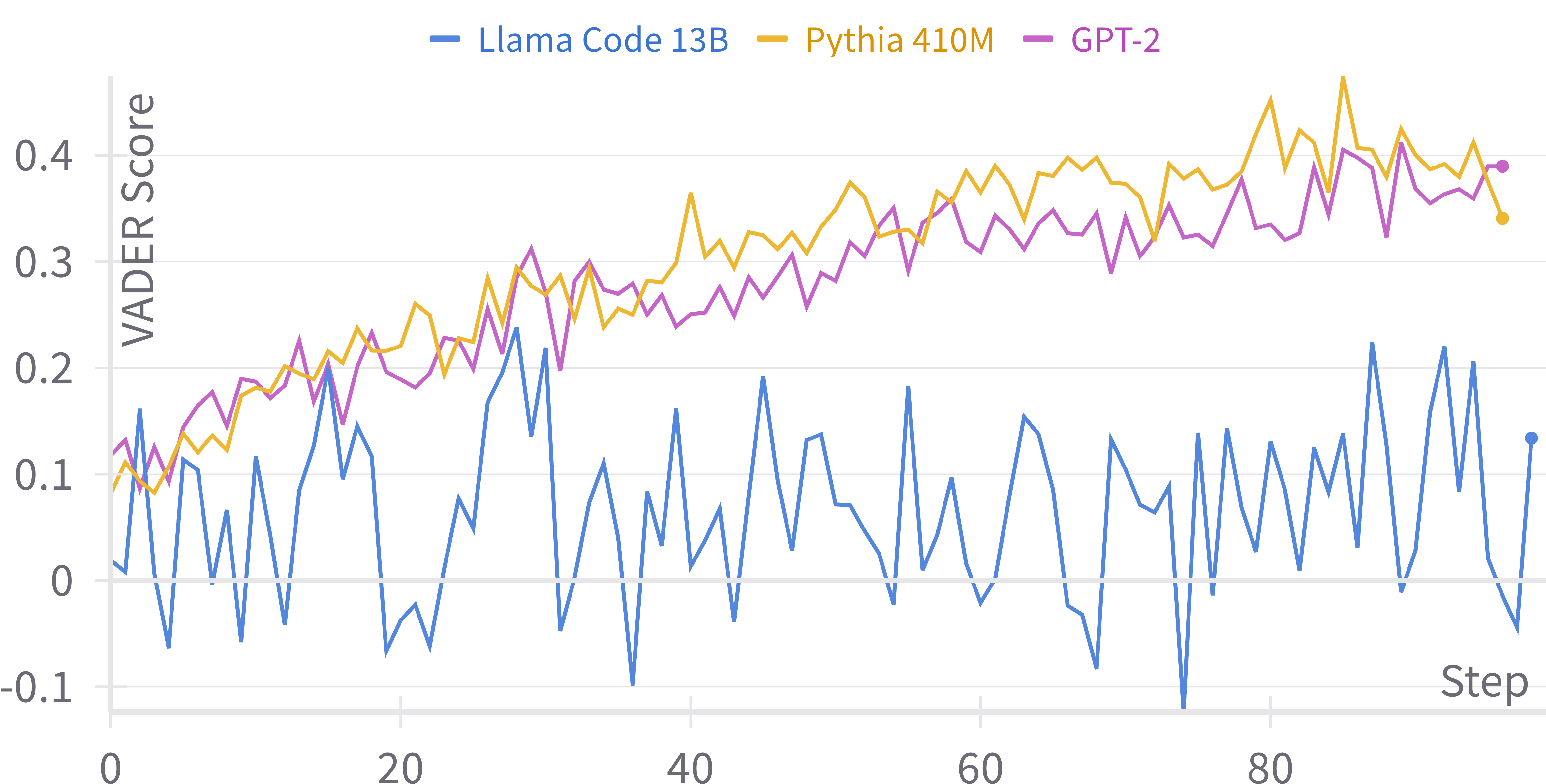}
    \caption{Comparison between three different base models, being trained with PPO for the sentiment alignment task, using \textbf{(top)} DistillBERT as a trained reward model, or \textbf{(bottom)} VADER as programmatic reward function.}
    \label{fig:imdb_model_comparison_bert_and_vader}
\end{figure}

\subsection{Sentiment Alignment Task}

In this first experiment, we isolate each modification that we have introduced to TRL to ascertain their individual impacts on the performance of the system for the sentiment alignment task. In \reffigure{fig:imdb_model_comparison_bert_and_vader} (top), GPT-2's training run is consistent with Hugging Face's original results \cite{HuggingFace_2022_sentiment}. We also find that, while Pythia 410M converges as expected, LLama Code 13B fails to improve, despite the model being otherwise capable of generating sensible reviews during inference. We hypothesize that this is due to an incompatbility between (1) the version of TRL we use, (2) 8-bit quantization, and (3) Llama-architecture models. GPT-2 and Pythia 410M did not use quantization.

Replacing the sentiment rewards obtained from the DistillBERT model with rewards calculated using the VADER algorithm, we find that the training runs in \reffigure{fig:imdb_model_comparison_bert_and_vader} (bottom) are consistent with those using DistillBERT, with GPT-2 and Pythia 410M steadily improving while LLama Code 13B shows no significant gains. We do, however, note an increased variance in rewards obtained during LLama Code 13B's training run with VADER rewards.

\begin{figure}[t]
    \centering
    \includegraphics[width=.68\linewidth]{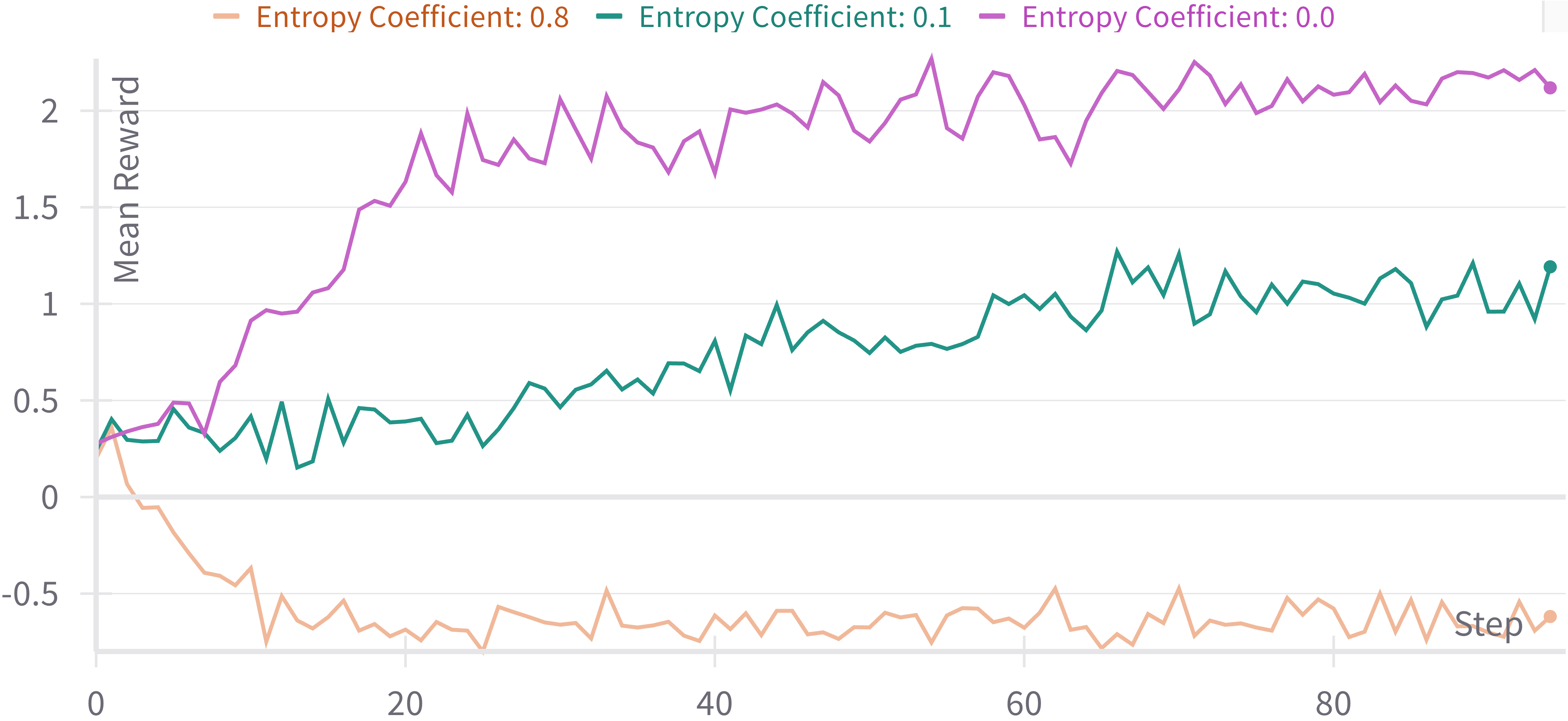}
    \includegraphics[width=.68\linewidth]{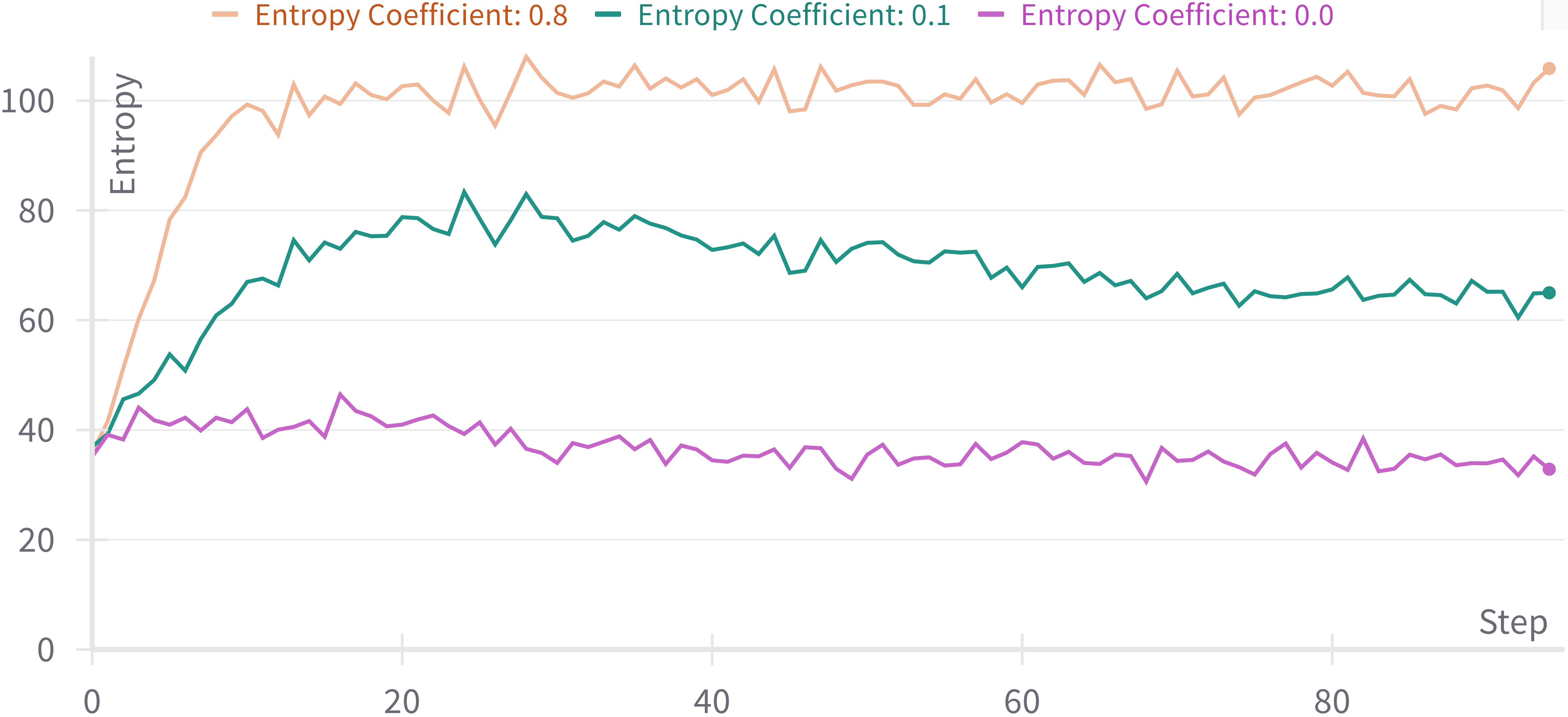}
    \caption{Comparison between three different values for the entropy regularization coefficient $\beta_{ENT}$ on the sentiment alignment task, using DistillBERT as a reward model.}
    \label{fig:imdb_entropies}
\end{figure}

\begin{figure}[t]
    \centering
    \includegraphics[width=.68\linewidth]{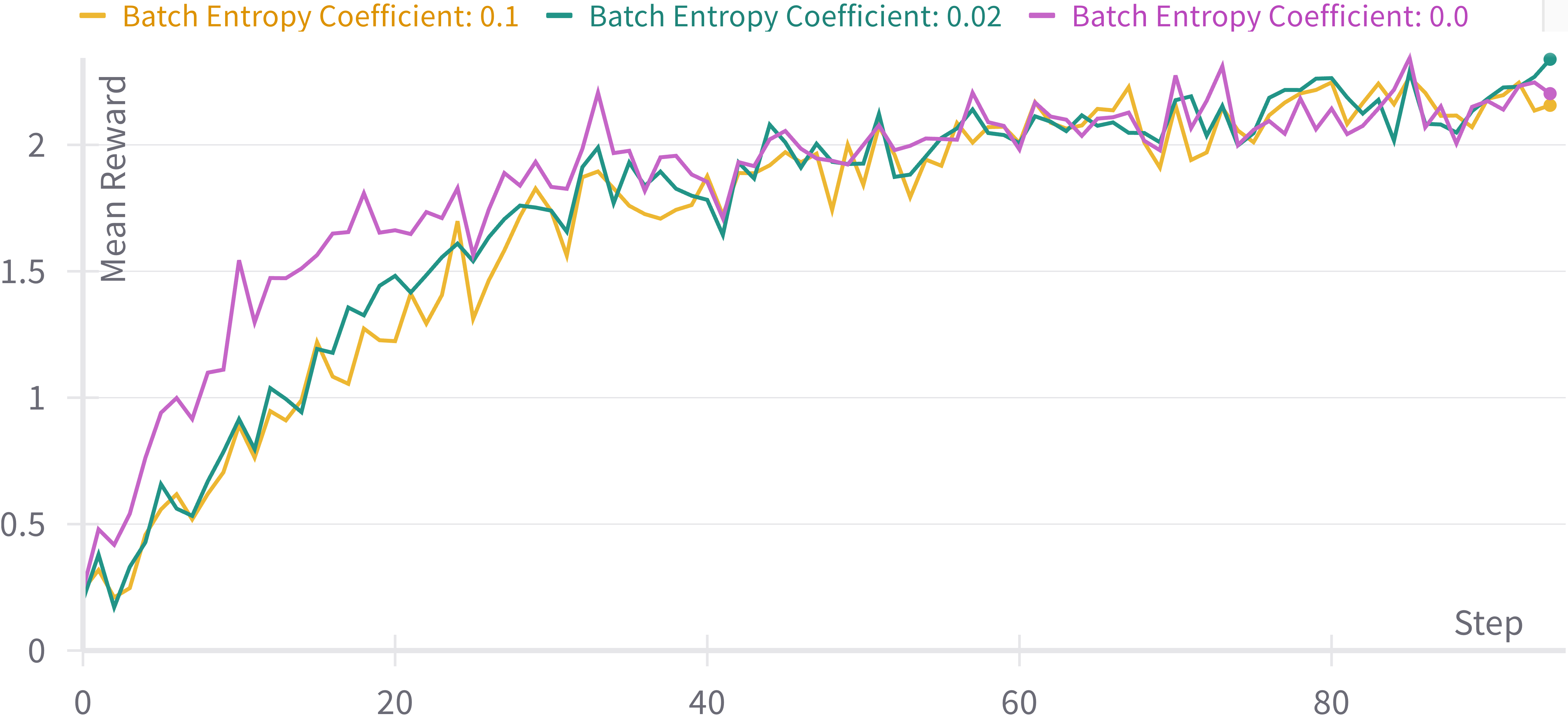}
    \includegraphics[width=.68\linewidth]{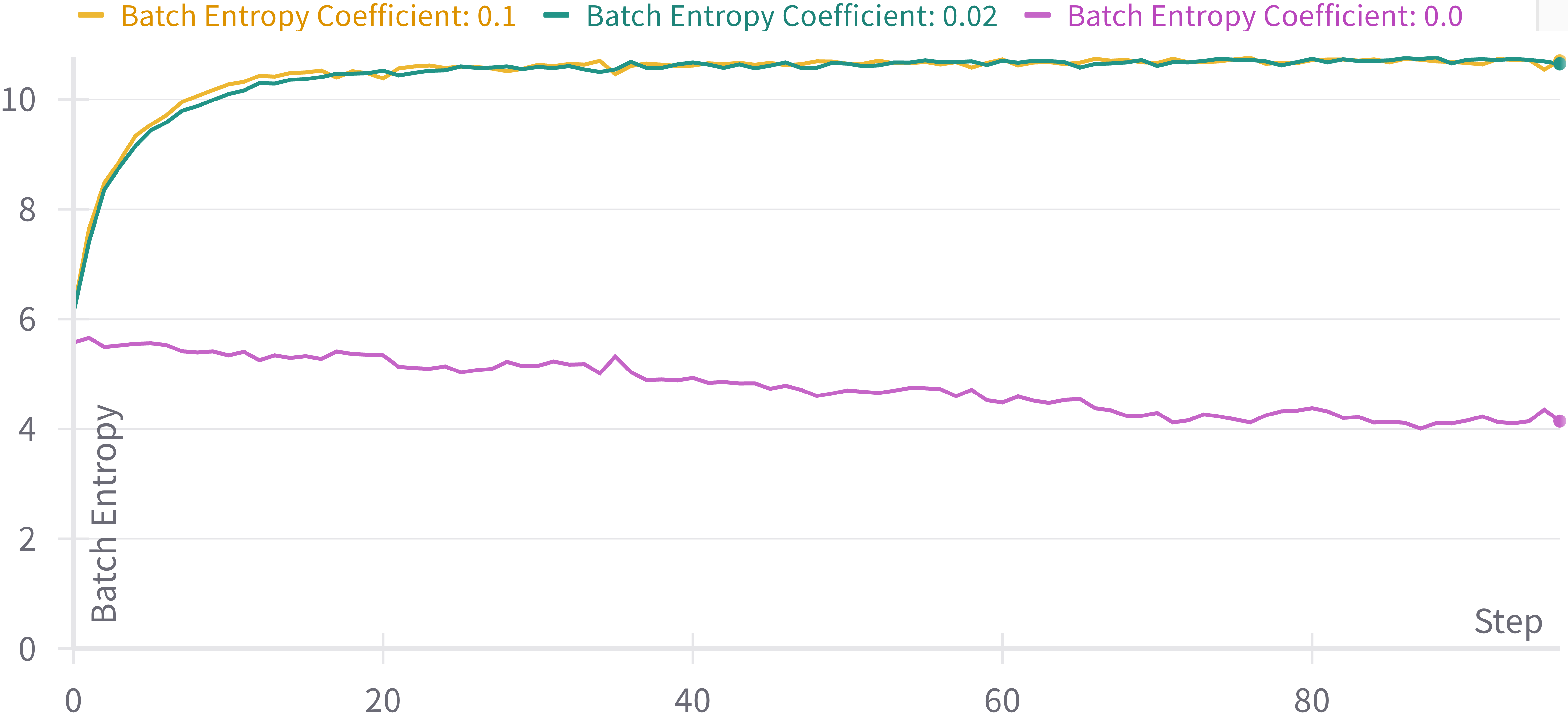}
    \vspace{-4pt}
    \caption{Comparison between training with PPO using three different values for $\beta_{BENT}$ on the sentiment alignment task, using DistillBERT as a reward model.}
    \label{fig:imdb_batch_entropies}
\end{figure}

\reffigure{fig:imdb_entropies} shows that raising the entropy regularization coefficient $\beta_{ENT}$ produces policies with higher entropy levels in their distributions over actions, as intended. However, in terms of rewards, this appears to lead to weaker policies. In contrast, when we raise the coefficient for our novel \textit{batch}-entropy regularization variant (\reffigure{fig:imdb_batch_entropies}), we can produce policies with higher levels of batch-entropy (note that these numbers are not directly comparable to regular entropy number), with no substantial detriment to rewards that the models converge to. We cannot rule out that similar results might be possible with the standard entropy regularization, but this would require at least a more thorough hyperparameter sweep.

%The entropy loss appears to correctly encourage a more uniformly distributed stochastic policy. \reffigure{fig:imdb_entropies} shows that larger coefficients for the entropy loss lead to a more uniform policy, however, the coefficient values which we tested are detrimental to the model's performance on the sentiment alignment task.

%\reffigure{fig:imdb_batch_entropies} is consistent with the correct implementation of the batch entropy loss. The batch entropy loss coefficients that we tested also seem to have a less detrimental effect on performance then the entropy coefficients did whilst still effectively encouraging the exploration of different actions across different states.

\begin{figure}[t]
    \centering
    \includegraphics[width=.7\linewidth]{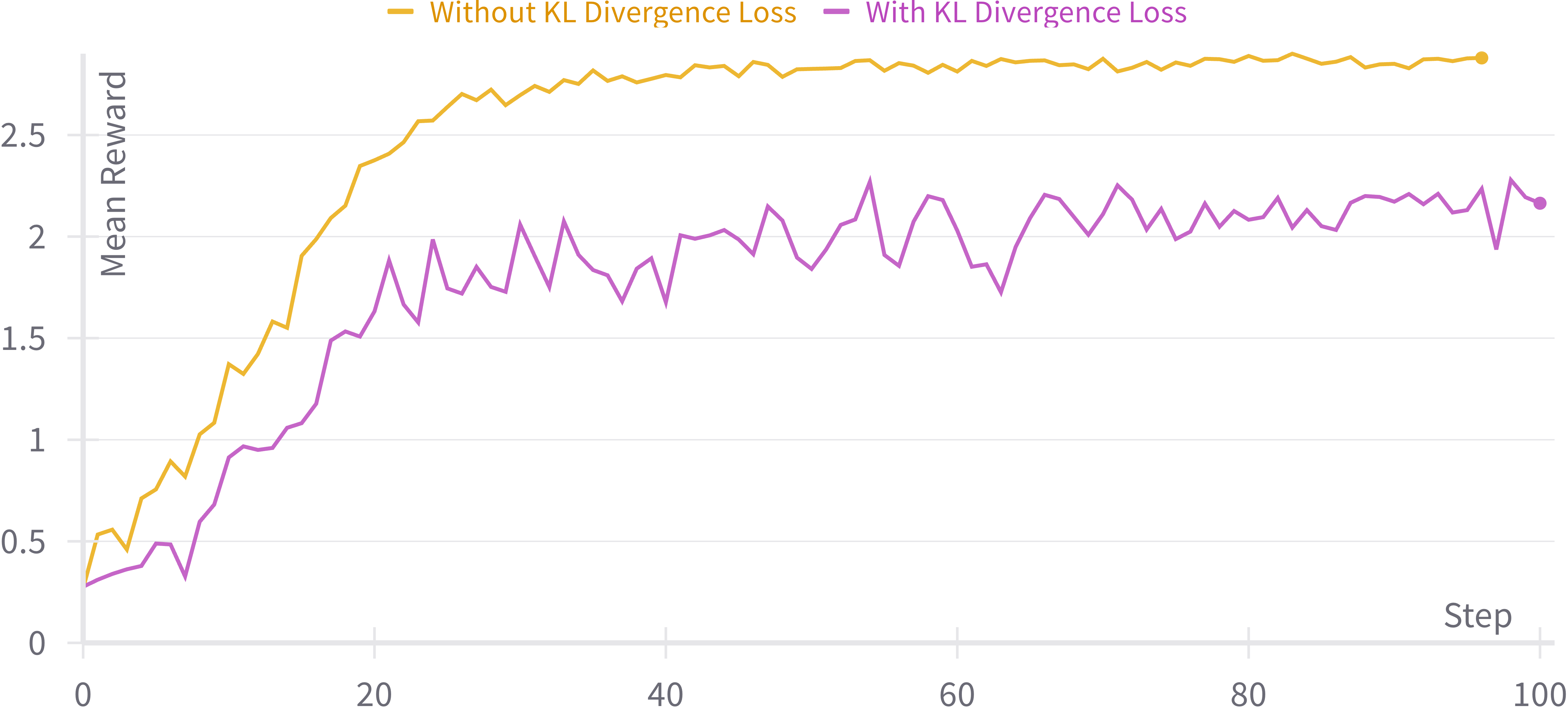}
    \caption{Comparison between training using PPO with and without the KL penalty on the sentiment alignment task, using DistillBERT as a reward model.}
    \label{fig:imdb_kl}
\end{figure}

Removing the KL divergence penalty (see \reffigure{fig:imdb_kl}) improved both the convergence rate as well as the final performance. However, we cannot rule out the possibility that allowing the model to drift further from the pretrained model may have decreased the overall natural language quality of the outputs (e.g., in terms of style or grammar) whilst improving in terms of positive sentiment.

\subsection{Arithmetic Task}

In this experiment, we train a GPT-2-based model from scratch, tailored to handle arithmetic expressions. The model is configured with a context size of 64 tokens and utilizes a new word-piece tokenizer. The tokenizer's vocabulary consists of integers from 0 to 45, and the symbols `+' and `='.

\begin{figure}[t]
    \vspace{-6pt}
    \centering
    \includegraphics[width=.48\linewidth]{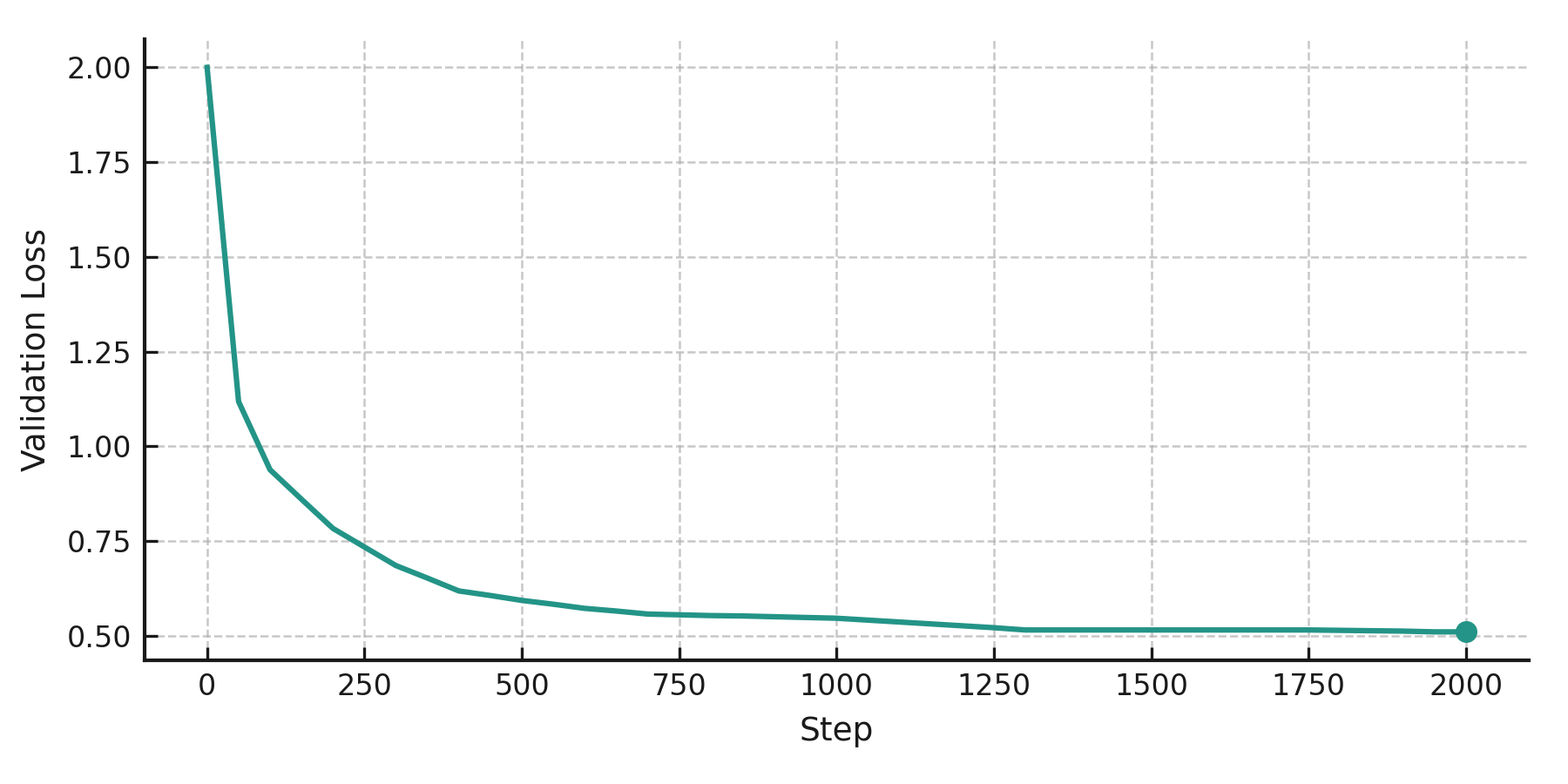}
    \includegraphics[width=.48\linewidth]{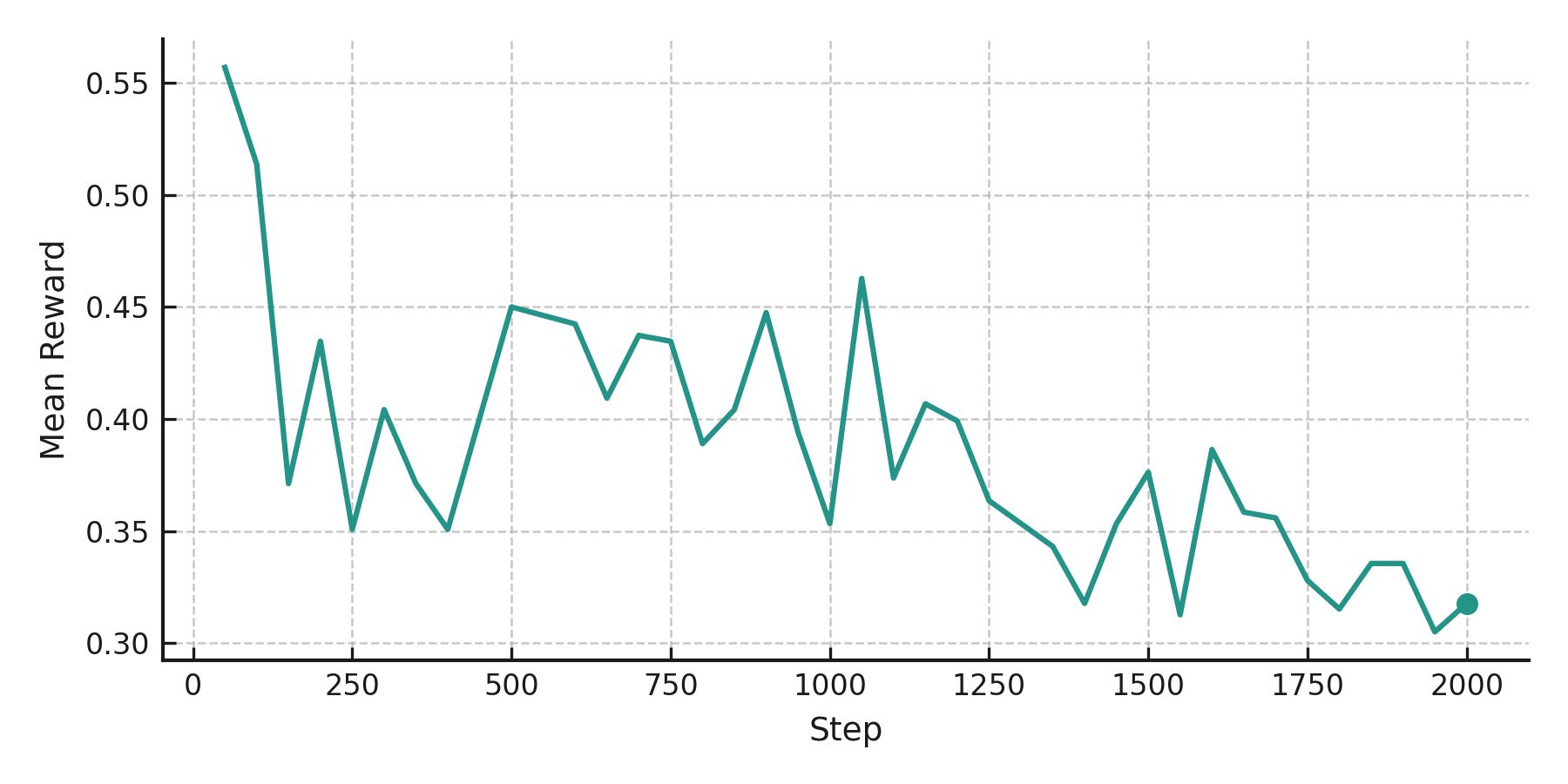}
    \caption{Validation Loss (left) and Mean Reward (right) during supervised MLM training on the arithmetic task.}
    \label{fig:math_MLM}
    \vspace{-6pt}
\end{figure}

\reffigure{fig:math_MLM} illustrates training under a conventional masked language modeling (MLM) objective. While the validation loss appears to converge, suggesting learning under the MLM objective, the reward deteriorates over time. The model learns to replicate approximately the correct structure, but fails to understand the mathematical semantics of the task.

\begin{figure}[ht!]
    \centering
    \includegraphics[width=.7\linewidth]{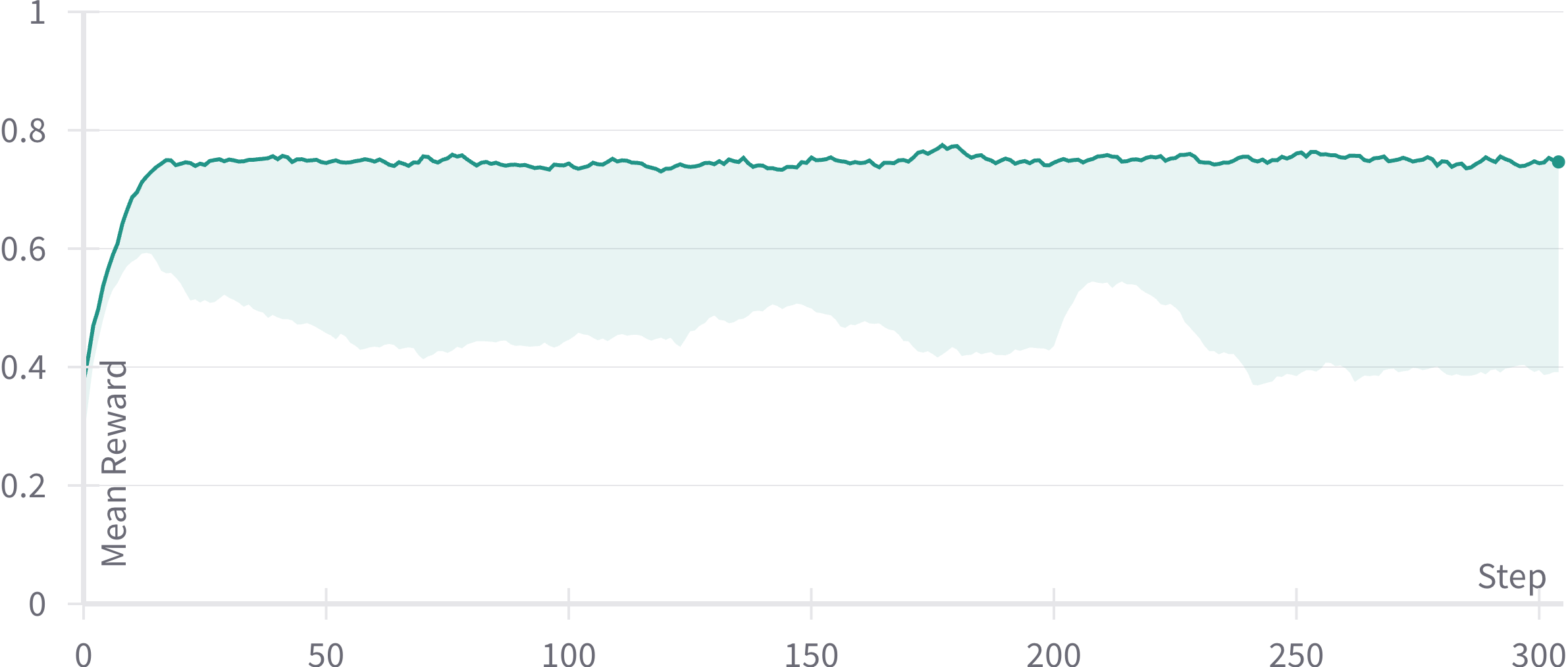}
    \caption{Maximum and (smoothed) minimum rewards across six PPO training runs on the arithmetic task without entropy or batch-entropy regularization.}
    \label{fig:math_baseline}
\end{figure}

\begin{figure}[t]
    \centering
    \includegraphics[width=.7\linewidth]{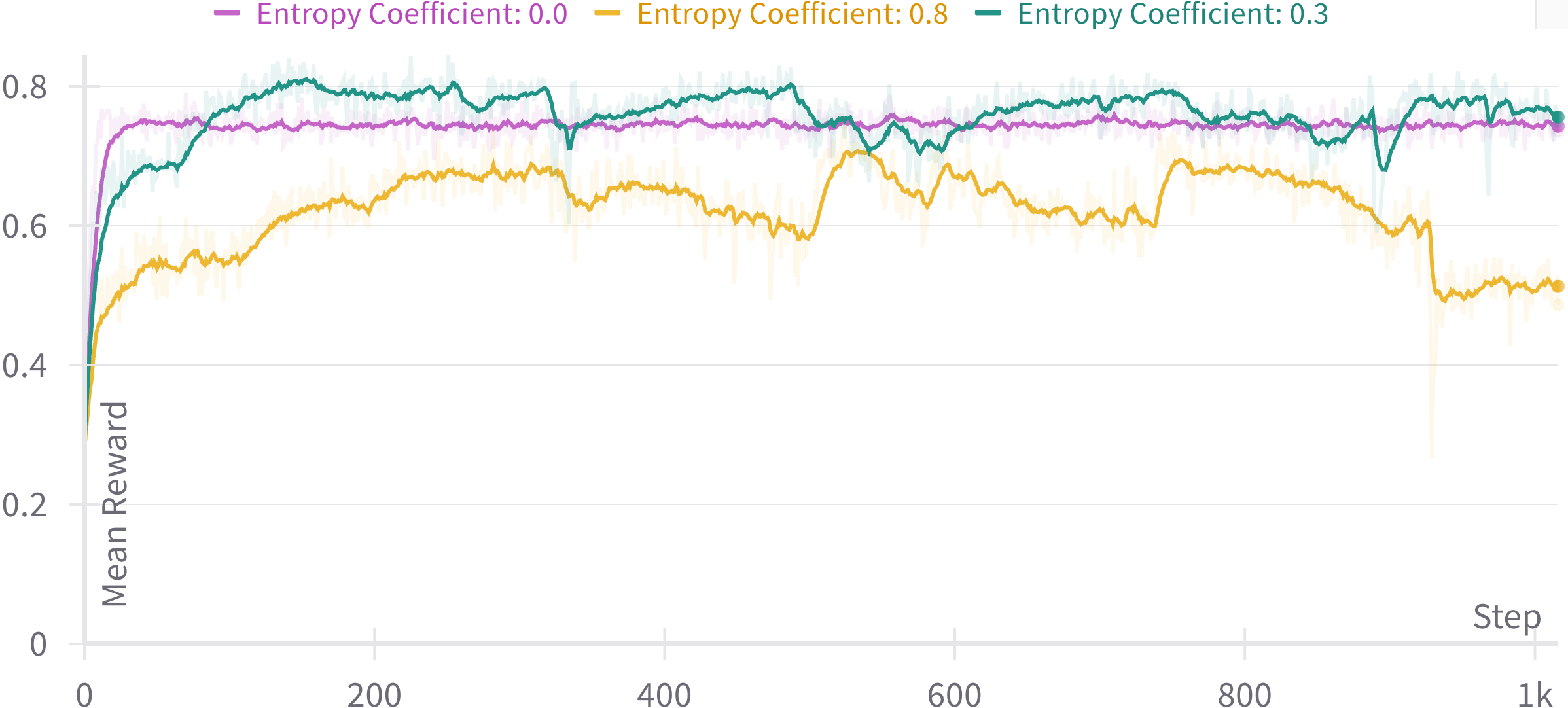}
    \includegraphics[width=.7\linewidth]{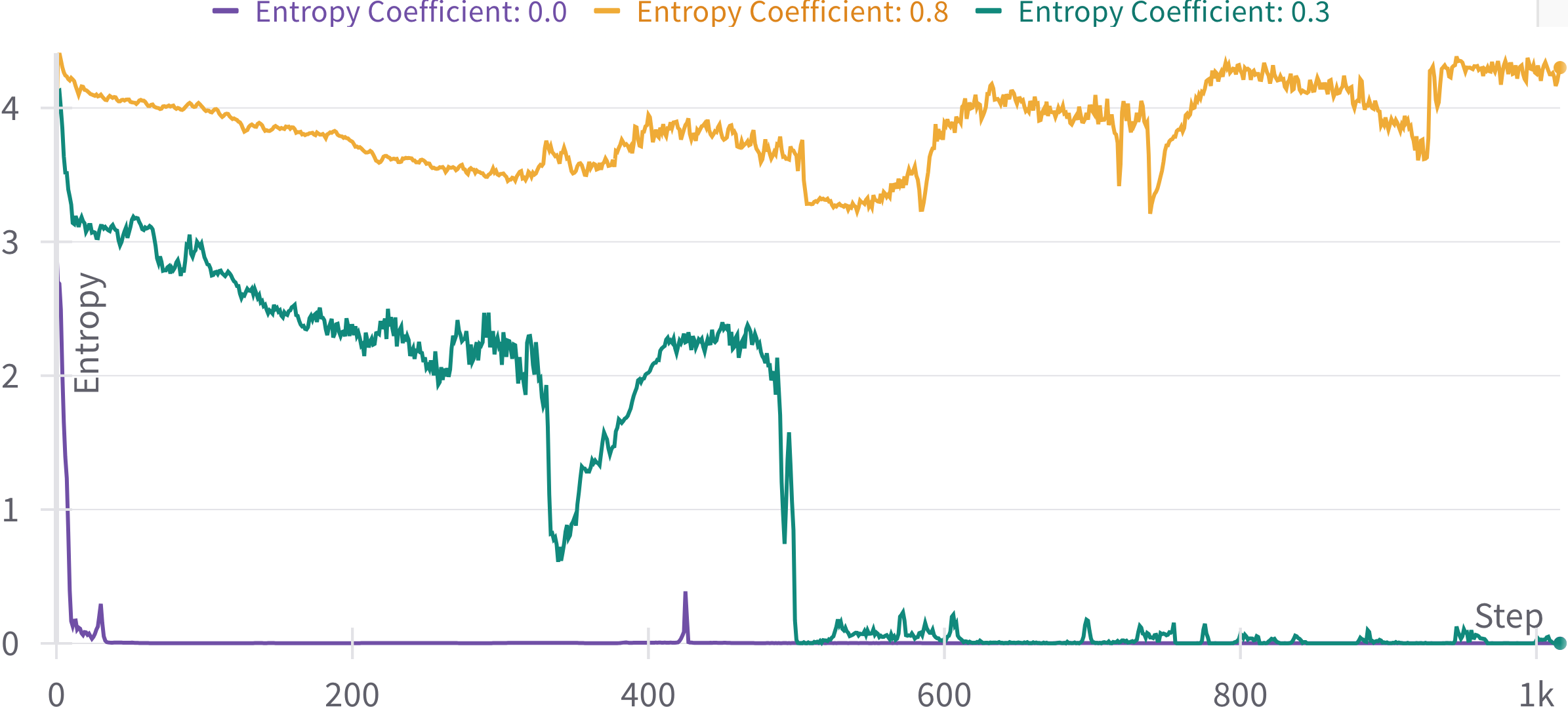}
    \caption{Comparison between training with PPO using three different values for $\beta_{ENT}$ on the arithmetic task. Rewards were smoothed to improve readability.}
    \label{fig:math_entropies}
\end{figure}

\begin{figure}[t]
    \centering
    \includegraphics[width=.7\linewidth]{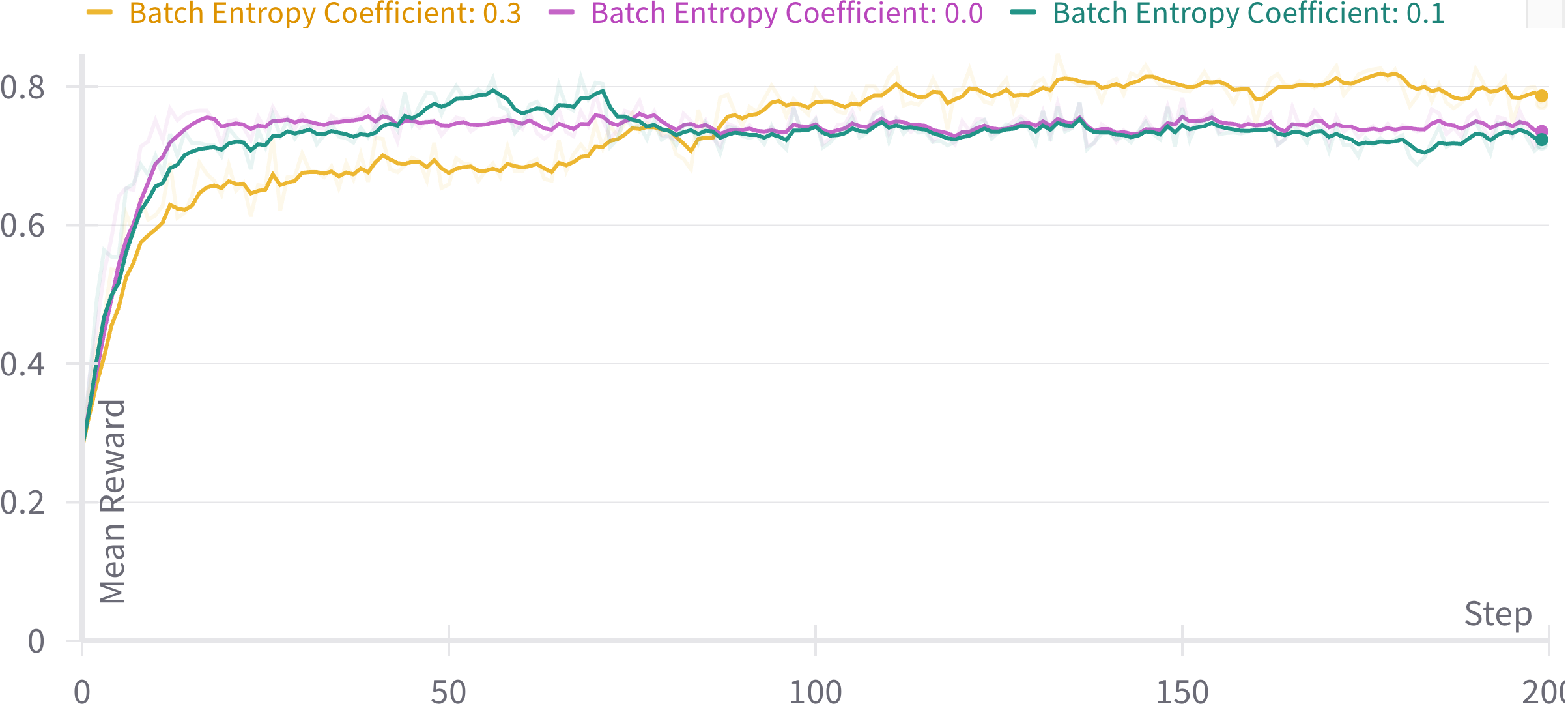}
    \includegraphics[width=.7\linewidth]{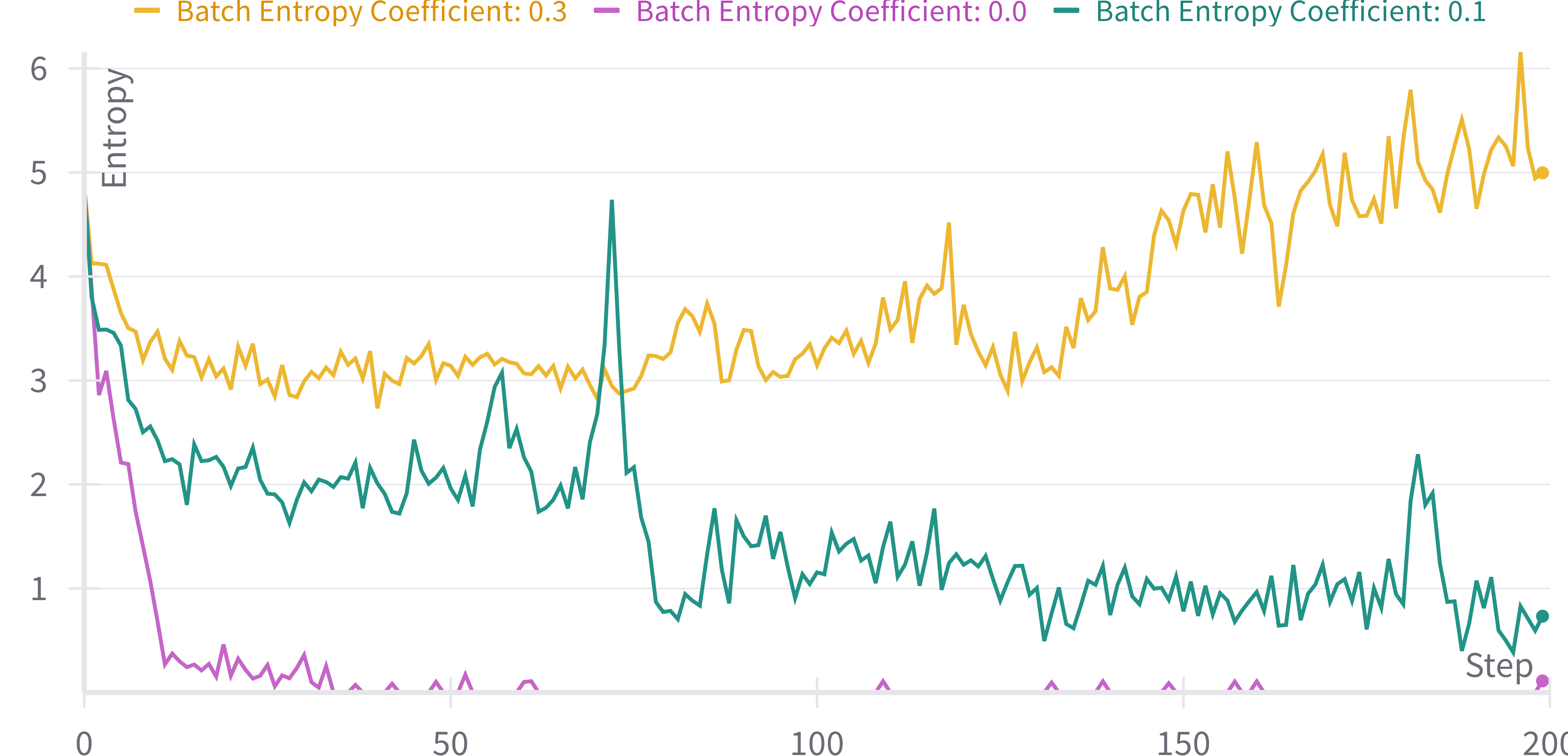}
    \caption{Comparison between training with PPO using three different values for $\beta_{BENT}$ on the arithmetic task. Rewards were smoothed to improve readability.}
    \label{fig:math_batch_entropies}
\end{figure}

As pre-training was largely ineffective, we start PPO training for the arithmetic task with an untrained model and no KL divergence penalty. \reffigure{fig:math_baseline} shows training using PPO to be more effective. The model quickly learns to output valid expressions and makes increasingly educated guesses toward the fully simplified expression, though it does not converge to a perfect solution. Figs. \ref{fig:math_entropies} and \ref{fig:math_batch_entropies} show that without entropy or batch-entropy regularization, the entropy rapidly collapses, and the model converges on a naive policy which generates \(23\) regardless of the prompt it is given. This is a notable local optimum: it is the (rounded) mean of the population of problems we can generate in this task, as $\mathbb{E}\left[ X1 + X_2 + X_3 + X_4 + X_5 \right] = 22.5$ when $X_i \sim \{0, 1, \dots, 9 \}$. With \(\beta_\text{ENT} = 0.3\) or \(\beta_\text{BENT} = 0.3\), the entropy collapse can be delayed and the model's performance can exceed that of the naive policy. However, increasing $\beta_\text{ENT}$ also appears to destabilize training.

\subsection{Ludii Game Synthesis Task}

The grammar of Ludii's DSL is complex enough that an untrained policy will face a flat reward landscape. This sets it apart from the arithmetic task, where it was feasible to start PPO training from an untrained model. In this task it is instead critical to first pre-train a minimally proficient model using a supervised MLM objective. We define a GPT-2 model with a custom tokenizer made up of all possible ludemes and primitives in the Ludii DSL. The GPT-2 variant was trained to convergence on the Ludii fill-in-the-middle dataset. However, despite efforts to simplify the tasks' representation using string masking (masking arbitrary strings such as names of games and pieces) and a custom tokenizer for the Ludii DSL, the model consistently failed to obtain a non-zero reward. 
%\textcolor{red}{(I guess the dataset is just too small to train a model from scratch.)} This presents a significant challenge, as the lack of reward variance creates a flat reward landscape, making it impossible for reward-based training to effectively guide the learning process.

%[!t] is flexible [H] isn't [!ht] is a mix

Fine-tuning Pythia 410M was more effective. Training this model to convergence on the training split of the Ludii dataset led to a mean reward above \(0.9\) out of \(1\) for both the training and validation splits. This is largely possible because the fill-in-the-middle dataset overrepresents smaller parentheticals. Filtering the dataset to only games where at least $20\%$ of the game description has been masked, the model's validation reward averages around \(0.3\), offering ample space for improvement with reward-based training. While the Llama Code 13B model also achieved comparable pre-training performance, we were forced to exclude it from further reward-based training since \reffigure{fig:imdb_model_comparison_bert_and_vader} suggests that Llama Code 13B is incompatible with the version of TRL that we used.

\begin{figure}[ht!]
    \centering
    \includegraphics[width=.7\linewidth]{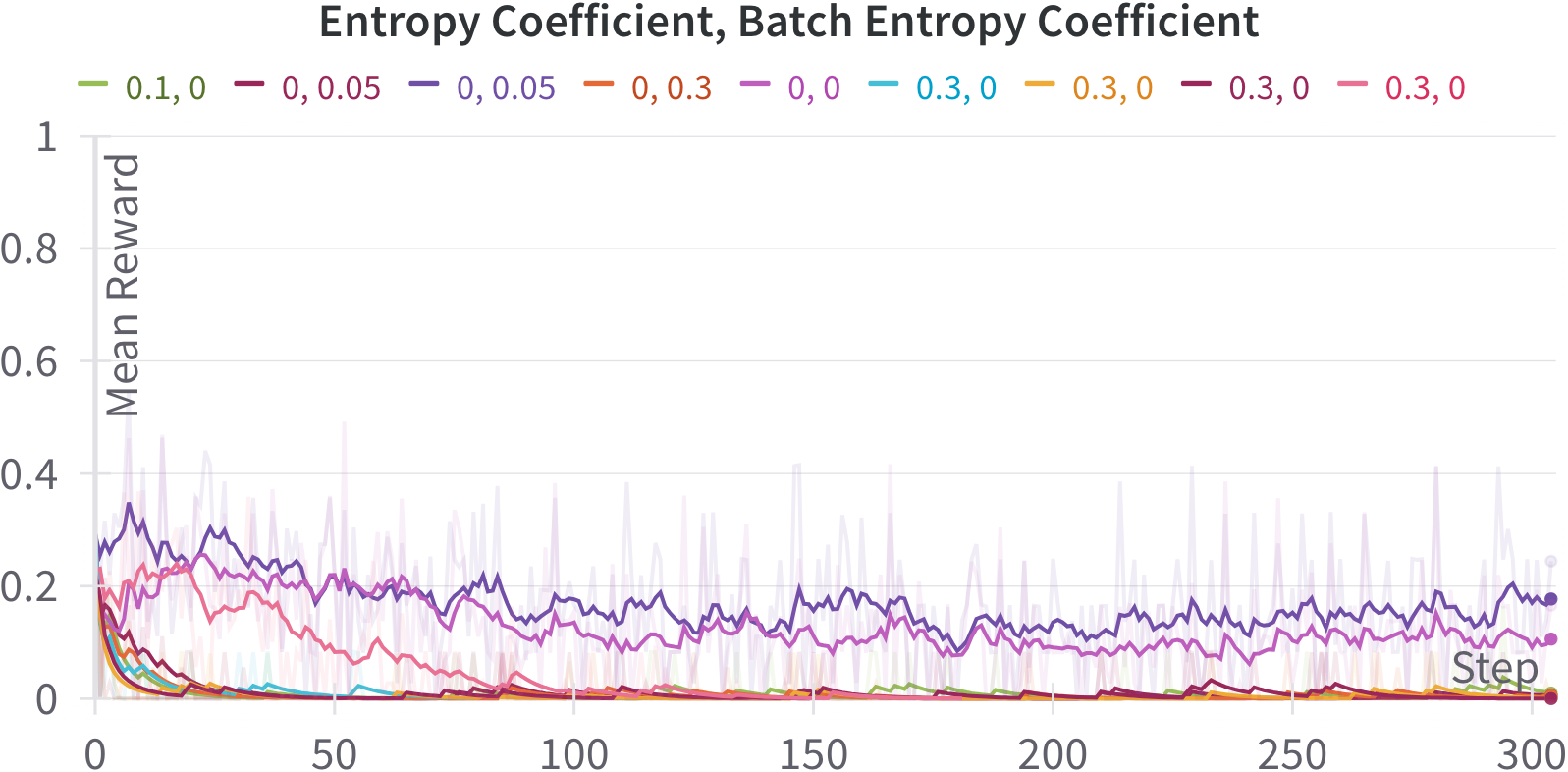}
    \caption{Comparison between training with PPO using 11 different values for the $\beta_{\text{ENT}}$ and $\beta_{\text{BENT}}$ coefficients on the Ludii game synthesis task. Runs are labeled as \(\beta_\text{ENT}, \beta_\text{BENT}\). Rewards were smoothed to improve readability.}
    \label{fig:ludii_individual_trajectories}
\end{figure}

None of the 11 PPO training runs that we conducted were able to improve the policy on the Ludii game synthesis task (see \reffigure{fig:ludii_individual_trajectories}). Runs with larger entropy and batch-entropy regularization coefficients also appear to diverge more quickly. We tried reintroducing the KL divergence penalty, but noted no difference in behavior. The fact that neither form of entropy regularization improved the performance of PPO on this task suggests that the observed instability may not be (solely) attributed to a lack of exploration. One possibility is that this task is too complex and requires a substantially larger model than Pythia 410M. It is also possible that PPO, or reinforcement learning more generally, may not be sufficiently stable for LLM training tasks that go beyond RLHF-style alignment, where minor parameter adjustments are made to encourage or discourage previously acquired capabilities, possibly due to a loss of plasticity. \cite{Lyle_2023_Understanding,Dohare_2024_LossOfPlasticity}.

\section{Related Work}

Prior work on using direct RL (as opposed to RL with trained reward models) to fine-tune LLMs for formal language tasks tends to focus on settings for which ample training data is available, and pre-trained models are already highly capable, such as commonly used programming languages \cite{le2022coderl,shojaee2023execution}. Outside of pure RL methods, researchers have also looked towards novel inference strategies to address similar shortcomings to those considered in this paper. Examples include iterative procedures for program synthesis that feed results or error reports from unit tests back into an LLM via additional prompts \cite{Liventsev2023AutonomousCodingLLMs}, adding lookahead search on top of RL to improve the math abilities of LLMs \cite{uesato2022solving}, and combining an LLM with evolutionary search for Ludii-based game synthesis \cite{Todd_2024_GAVEL}.

\section{Conclusions \& Future Work}

This paper explores the feasibility of using direct Reinforcement Learning (RL) with programmed reward functions (as opposed to the more common trained reward models used in Reinforcement Learning from Human Feedback) to fine-tune LLMs for formal-language tasks which the model was not exposed to during pre-training.

Our first experiments replicate prior work on sentiment analysis \cite{HuggingFace_2022_sentiment} and validate the correctness of our TRL-based implementation \cite{vonwerra2022trl}. We then designed an arithmetic task that could not be effectively learned through supervised learning alone. RL-based training proved to be more effective; however, without entropy regularization, the model consistently converged to a naive local optimum. Both classical entropy regularization and our novel form of batch-entropy regularization improved upon this local optimum. While, theoretical reasoning and our empirical results suggest that batch-entropy regularization provides greater stability, a comprehensive hyperparameter sweep would be needed to confirm this observation. Our final experiments found that reward-based training of GPT-2 and Pythia 410M for the complex task of generating board games in Ludii's game description language was unstable.

Our initial observation that PPO is effective at model alignment, such as encouraging a pre-trained model to write exclusively positive reviews, is consistent with the literature. However, we found that this performance does not generalize to unseen tasks, like learning to design board games, or even simply summing numbers. Since PPO is a state-of-the-art RL training algorithm, our findings highlight the need for fundamental improvements in RL training algorithms for reward-based training of LLMs. Potential avenues worth exploring include simpler methods like RLOO \cite{Ahmadian_2024_BackToBasics} and incorporating better domain-specific inductive biases, such as task-specific positional encodings. For complex tasks, such as game synthesis, it is also plausible that substantially larger models or more computationally expensive search algorithms \cite{uesato2022solving,Todd_2024_GAVEL} are required. Nevertheless, exploring the limits of improving RL training before resorting to such resource-intensive methods remains a compelling area of investigation.

\begin{credits}
\subsubsection{\ackname} We thank Aki H{\"a}rm{\"a} for feedback on an early draft of this work.
\subsubsection{\discintname}
The authors have no competing interests to declare that are relevant to the content of this article.
\end{credits}

\bibliographystyle{splncs04}
\bibliography{references,Dennis-Soemers-Bib}

\end{document}